\documentclass[10pt, conference, compsocconf]{IEEEtran}
\pdfoutput=1
\usepackage{cite}
\usepackage{times}
\usepackage{color,bm}

\usepackage{graphicx}
\usepackage{subfig}
\usepackage{bm,latexsym,mathrsfs}
\usepackage{amsthm,amsmath,amsfonts}
\usepackage{MnSymbol }
\usepackage{tikz,soul}
\usepackage{placeins}
\usepackage{url}
\usepackage{hyperref}

\newcommand{\E}{{\mathbb{E}}}

\newcommand{\Var}{{\mathbb{V}\text{ar}}}
\newcommand{\Cov}{{\mathbb{C}\text{ov}}}

\newcommand*{\mean}[1]{\mathbb{E} \left[{#1}\right]}

\newcommand\independent{\protect\mathpalette{\protect\independenT}{\perp}}
\def\independenT#1#2{\mathrel{\rlap{$#1#2$}\mkern2mu{#1#2}}}

\newtheorem{thm}{Theorem}

\newtheorem{lem}{Lemma}
\theoremstyle{definition}
\newtheorem{defn}{Definition}

\begin{document}

\title{Hierarchical Multinomial-Dirichlet model for the estimation of \\conditional probability tables}

\author{\IEEEauthorblockN{Laura Azzimonti}
\IEEEauthorblockA{IDSIA - SUPSI/USI\\
Manno, Switzerland\\
laura@idsia.ch}
\and
\IEEEauthorblockN{Giorgio Corani}
\IEEEauthorblockA{IDSIA - SUPSI/USI\\
Manno, Switzerland\\
giorgio@idsia.ch}
\and
\IEEEauthorblockN{Marco Zaffalon}
\IEEEauthorblockA{IDSIA - SUPSI/USI\\
Manno, Switzerland\\
zaffalon@idsia.ch}
}

\maketitle
\begin{abstract}
We  present a novel approach for estimating conditional probability tables, based on a joint, rather than independent, estimate of the conditional distributions belonging to the same table. We derive exact analytical expressions for the estimators and we analyse their properties both analytically and via simulation.
We then apply this method to the estimation of parameters in a Bayesian network.  
Given the structure of the network, the proposed approach better estimates the joint distribution and significantly improves the classification performance with respect to traditional approaches. 
\end{abstract}


\section{INTRODUCTION}

A Bayesian network is a probabilistic model constituted by a directed acyclic graph (DAG) and a set of 
\textit{conditional probability tables} (CPTs), one for each node.
The  CPT of node $X$ contains the conditional probability distributions of $X$ given each possible configuration of its parents.
Usually all variables 
are discrete and 
the conditional distributions are estimated 
adopting a Multinomial-Dirichlet model, where the
Dirichlet prior is characterised by the vector of hyper-parameters $\bm \alpha$.
Yet, Bayesian estimation of multinomials is sensitive to the choice of 
$\bm \alpha$ and 
inappropriate values 
cause the estimator to perform poorly \cite{hausser2009entropy}.
Mixtures of Dirichlet distributions have been recommended
both in statistics \cite{casella2009assessing,good1987robustness}  
and in machine learning \cite{nemenman2002entropy}
in order to obtain more robust estimates.
 Yet, mixtures of Dirichlet distributions are computationally expensive; this prevents them from being widely adopted.
Another difficulty encountered in CPT estimation is the presence of rare events.
Assuming that all variables have cardinality $k$ and that the number of parents is $q$,
we need to estimate $k^q$ conditional distributions, one for each
joint configuration of the parent variables. Frequently one or more of such configurations are rarely observed in the data, making their estimation challenging.

We propose to estimate the conditional distributions by 
adopting a novel approach, based on a hierarchical Multinomial-Dirichlet model.
This model has two main characteristics.
First, the prior of each conditional distribution is 
constituted by 
a mixture of Dirichlet distributions
with {parameter} $\bm \alpha$; the mixing is attained by treating
$\bm \alpha$ as a random variable with its own prior and posterior distribution.
By estimating from data
the posterior distribution of $\bm \alpha$, we need not to fix its value \emph{a priori}.
Instead we give more weight
to the values of $\bm \alpha$  that are more likely
given the data.
Secondly, the model is hierarchical since it assumes that
the conditional distributions within the same CPT (but referring to different joint configurations of the parents) are drawn from the same mixture.
The hierarchical model 
\textit{jointly} estimates all the conditional distributions 
of the same CPT, called \emph{columns} of the CPT.
The joint estimates generate information flow between  different columns of the CPT; thus the hierarchical model exploits the parameters learned for data-rich columns to improve the estimates of the parameters of data-poor columns.
This is called \textit{borrowing statistical strength} \cite[Sec 6.3.3.2]{murphy2012machine}
and it is well-known within the literature of hierarchical models \cite{gelman2013bayesian}.
Also the literature of Bayesian networks acknowledges  \cite[Sec.17.5.4]{koller2009probabilistic} that introducing dependencies between  columns 
of the same CPT could improve the estimates,  especially when dealing with sparse data.
However, as far as we known, this work is the first practical application of joint estimation of the columns of the same CPT.

To tackle the problem of computational complexity we adopt a variational inference approach. Namely, we compute a factorised approximation of the posterior distribution that is highly efficient. 
Variational inference appears particularly well suited for hierarchical models; for instance the inference of Latent Dirichlet Allocation \cite{LDA} 
is based on variational Bayes.
By extensive experiments, we show that our novel approach considerably improves
parameter estimation compared to the traditional approaches based on Multinomial-Dirichlet model. The experiments show  large gains especially when dealing with small samples, while with large samples the effect of the prior vanishes as expected.

The paper is organised as follows. Section \ref{sec:param_indep} introduces the novel hierarchical model.
Section \ref{sec:hier_mod} provides an analytical study of the resulting estimator, proving that the novel hierarchical approach provides lower estimation error than the traditional approaches, 
under some mild assumptions on the generative model.
Section \ref{sec:exp} presents some simulation studies showing that, given the same network structure,
hierarchical estimation yields both a better fit of the joint distribution and a consistent improvement in classification performance,
with respect to the traditional estimation under parameter independence. 
Section \ref{sec:concl} reports some concluding remarks.

 \section{ESTIMATION UNDER  {MULTINOMIAL-DIRICHLET MODEL}}
\label{sec:param_indep}
We want to induce a Bayesian network over the set of random variables $\bm{X}= \{X_1,\ldots,X_I\}$.
We assume that each variable $X_i \in \bm{X}$ is discrete and has $r_i$ possible values in  the set
$\mathcal{X}_i$. 
The parents of $X_i$ are denoted by $\mathrm{Pa}_i$ and they have $q_i$ possible joint states collected in the set
$\mathcal{P}_i$. \\
We denote by $\theta_{x|\mathrm{pa}}$ the probability of $X_i$ being in state $x\in \mathcal{X}_i$ when its parent set is in state $\mathrm{pa}\in\mathcal{P}_i$, i.e., $\theta_{x|\mathrm{pa}} = p(X_i=x |  \mathrm{Pa}_i=\mathrm{pa}) > 0$.
 We denote by $\bm{\theta}_{X_i|\mathrm{pa}}$ the parameters of the conditional distribution of $X_i$ given $\mathrm{Pa}_i=\mathrm{pa}$.
A common assumption \cite[Sec.17]{koller2009probabilistic} is that $\bm{\theta}_{X_i|\mathrm{pa}}$ is generated from a Dirichlet distribution with known parameters. 
 The collection of the conditional probability distributions $\bm{\theta}_{X_i} = (\bm{\theta}_{X_i|\mathrm{pa}_1},\ldots, \bm{\theta}_{X_i|\mathrm{pa}_{q_i}})$ constitutes  the \textit{conditional probability table} (CPT)  of $X_i$.
 Each  vector of type $\bm{\theta}_{X_i|\mathrm{pa}}$, with $\mathrm{pa}\in \mathcal{P}_i$, is a \textit{column} of the CPT.

The  assumption of  \textit{local parameter independence} \cite[Sec. 17]{koller2009probabilistic} 
allows to estimate each  parameter vector $\bm{\theta}_{X_i|\mathrm{pa}}$ independently of the other parameter vectors.
The assumed generative model, $\forall i \in 1,\ldots, I$, is:
\begin{align*}
p(\bm{\theta}_{X_i|\mathrm{pa}})&=\mathrm{Dir} (s \bm{\alpha}) &\mathrm{pa}\in \mathcal{P}_i, \nonumber \\
p(X_i|\mathrm{Pa}_i=\mathrm{pa},\bm{\theta}_{X_i|\mathrm{pa}}) &=\mathrm{Cat} (\bm{\theta}_{X_i|\mathrm{pa}}) &\mathrm{pa}\in \mathcal{P}_i,
\end{align*}
where $s\in \mathbb{R}$ denotes the prior strength, also called \textit{equivalent sample size}, and $\bm{\alpha}\in \mathbf{R}^{r_i}$ is a parameter vector such that $\sum_{x\in\mathcal{X}}\alpha_x=1$.
The most common choice is to set 
$\alpha_{x}=1/r_i$ and $s=1/q_i$, which is called BDeu prior \cite[Sec.17]{koller2009probabilistic}.

If there are no missing values in the data set $D$, the posterior expected value of  {$\theta_{x_|\mathrm{pa}}$ is \cite{gelman2013bayesian}:
\begin{align*}
& E [\theta_{x_|\mathrm{pa}}] = \frac{n_{x,\mathrm{pa}}+s \alpha_{x}}{n_{\mathrm{pa}}+s}, 
\end{align*}
where $n_{x,\mathrm{pa}}$ is the number of observations in $D$ characterised by $X_i=x$ and $\mathrm{Pa}_i=\mathrm{pa}$, while $n_{\mathrm{pa}}=\sum_{x \in \mathcal{X}_i}n_{x,\mathrm{pa}}$. 

\section{HIERARCHICAL 
MODEL}
\label{sec:hier_mod}
The proposed hierarchical model
estimates the conditional probability tables by removing the local independence assumption.
In order to simplify the notation we present the model on a  node $X$ with a single parent $Y$. 
$X$ has $r$ states in the set 
$\mathcal{X}$, while $Y$ has  $q$ states in the set $\mathcal{Y}$.
Lastly, we denote by $n_{xy}$ the number of observations with $X=x$ and $Y=y$ and by $n_{y}=\sum_{x\in \mathcal{X}} n_{xy}$ the number of observations with $Y=y$, where $x \in \mathcal{X}$ and $y \in \mathcal{Y}$. 

As described in Section \ref{sec:param_indep}, $\bm{\theta}_{X_i|\mathrm{pa}}$, for $i=1,\ldots,I$ and
$\mathrm{pa}\in \mathcal{P}_i$, are usually assumed to be independently drawn from a Dirichlet distribution, with known parameter 
$\bm{\alpha}$. 
On the contrary, the hierarchical model treats $\bm{\alpha}$ as a hidden random  {vector}, thus making different columns of the CPT dependent. 
Specifically, we assume $\bm{\alpha}$ to be drawn from a higher-level Dirichlet distribution with hyper-parameter 
$\boldsymbol{\alpha}_0$.

We assume that $(x_{k},y_{k})$ for $k= 1, \ldots, n$ are $n$ $i.i.d.$ observations from the hierarchical Multinomial-Dirichlet model:
\vspace{-5pt}
\begin{align}
p(\boldsymbol{\alpha} | \boldsymbol{\alpha}_0) & = \text{Dirichlet}(\boldsymbol{\alpha}_0) \nonumber \\
p(\boldsymbol{\theta}_{X|y} | s, \boldsymbol{\alpha}) & = \text{Dirichlet}(s\boldsymbol{\alpha}) & y \in \mathcal{Y} \label{eq:hier}
\\
p(X| Y=y, \boldsymbol{\theta}_{X|y}) & = \text{Cat}(\boldsymbol{\theta}_{X|y}) & y \in \mathcal{Y} \nonumber 
\end{align}
where  
$s\in \mathbb{R}$ is the equivalent sample size, and $\boldsymbol{\alpha}_0\in \mathbb{R}^{r}$ is a vector of hyper-parameters.

\subsection{Posterior moments for $\boldsymbol{\theta}_{X|y}$}
We now study the hierarchical model, deriving 
an analytical expression for the posterior  {average} of ${\theta}_{x|y}$, which is the element $x$ of vector $\boldsymbol{\theta}_{X|y}$, and for the posterior covariance between ${\theta}_{x|y}$ and ${\theta}_{x'|y'}$.
To keep notation simple, in the following we will not write explicitly the conditioning with respect to the fixed parameters $s$ and $\boldsymbol{\alpha}_0$. We introduce the notation $\E^{D}\left[ \cdot \right]=\mean{\left.\cdot\right| D}$ to represent the posterior  {average} and $\Cov^{D}{\left(\cdot, \cdot \right)}=\Cov{\left(\cdot, \left.\cdot\right| D\right)}$ to represent the posterior covariance.
\begin{defn}
We define the pointwise estimator $\hat{\theta}_{x|y}$ for the parameter $\theta_{x|y}$ as its posterior average, i.e.,
$\hat{\theta}_{x|y}=\E^{D}\left[\theta_{x|y}\right]$, 
and the pointwise estimator $\hat{\alpha}_x$ for the element $x$ of the parameter vector  $\bm{\alpha}$ as its posterior average, i.e.,
$\hat{\alpha}_x=\E^{D}\left[{\alpha}_x\right].$
\end{defn}
\begin{thm}
Under model \eqref{eq:hier}, the posterior  {average} of $\theta_{x|y}$ is
\vspace{-2pt}
\begin{equation}
\hat{\theta}_{x|y}=\E^{D}\left[\theta_{x|y}\right]=\frac{n_{xy}+s\hat{\alpha}_{x}}{n_y+s},
\label{eq:mean}
\end{equation}
while the posterior covariance between $\theta_{x|y}$ and $\theta_{x'|y'}$ is
\begin{equation*}
\Cov^{D}{\!\left(\!\theta_{x|y},\! \theta_{x'|y'}\!\right)}\!\!=\! \!\delta_{y y'}\! \frac{\hat\theta_{x|y}\delta_{x x'}\!- \!\hat\theta_{x|y}\hat{ \theta}_{x'|y}}{n_y+s+1} + \frac{s^2\Cov^{D}\!\! \left(\!\alpha_x,\! \alpha_{x'}\!\right)}{C_{y y'}},
\label{eq:cov}
\end{equation*}
 where $C_{y y'}$ is defined as
 \begin{equation*}
 C_{y y'}=
 \left\{
 \begin{array}{ll}
(n_y+s)(n_{y'}+s) & \text{if } y\neq y'\\
(n_y+s)(n_y+s+1) & \text{if } y=y'.
 \end{array}
 \right.
 \label{eq:cost_denom}
 \end{equation*}
\label{thm:theta_mean}
\end{thm}
\vspace{-10pt}
The posterior average and posterior covariance of $\boldsymbol{\alpha}$ cannot be computed analytically. Some results concerning their expression and numerical computation, together with the complete proof of Theorem \ref{thm:theta_mean}, are detailed in Appendix.

Notice that the pointwise estimator $\hat{\theta}_{x|y}$ is a mixture of traditional Bayesian estimators obtained under (non-hierarchical) Multinomial-Dirichlet models with $\boldsymbol{\alpha}$ fixed, i.e, $\frac{n_{xy}+s\alpha_{x}}{n_y+s}$. Indeed, thanks to the linearity in $\alpha_{x}$, we obtain 
\begin{equation*}
\hat{\theta}_{x|y}=\frac{n_{xy}+s\hat{\alpha}_{x}}{n_y+s}=\int{\frac{n_{xy}+s\alpha_{x}}{n_y+s} p(\boldsymbol{\alpha}|D)d\boldsymbol{\alpha}}.
\end{equation*} 
This mixture gives more weight to the values of $\boldsymbol{\alpha}$ that are more likely given the observations.

\subsection{Properties of the estimator $\hat{\boldsymbol{\theta}}_{X|y}$}
\label{sec:MSE}
We study now 
the mean-squared error (MSE) of $\hat{\theta}_{x|y}$ and we compare it to the MSE of other traditional estimators. \\
In order to study the MSE of $\hat{\theta}_{x|y}$  we need to assume the generative model
\vspace{-2pt}
\begin{align}
p(\boldsymbol{\theta}_{X|y} | s, \tilde{\boldsymbol{\alpha}}) & = \text{Dirichlet}(s\tilde{\boldsymbol{\alpha}}) & y \in \mathcal{Y}, \nonumber\\
p(X| Y=y, \boldsymbol{\theta}_{X|y})& = \text{Cat}(\boldsymbol{\theta}_{X|y}) & y \in \mathcal{Y} , \label{eq:hier_generative}
\end{align}
\vspace{-2pt}
\!where $s$ and $\tilde{\boldsymbol{\alpha}}$ are the true underlying parameters. 
Moreover, since $\boldsymbol{\theta}_{X|y}$ is a random vector, we define the MSE for an estimator $\bar{\theta}_{x|y}$ of the single component 
${\theta}_{x|y}$ 
as
\begin{equation}
\text{MSE}\left(\bar{\theta}_{x|y} \right)=\E_{\theta}\left[\E_{n}\left[\left(\bar{\theta}_{x|y}-{\theta}_{x|y} \right)^2\right]\right],
\label{eq:MSE}
\end{equation}
where $\E_{\theta}\left[\cdot \right]$ and $\E_{n}\left[\cdot\right]$ represent respectively the expected value with respect to ${\theta_{x|y}}$ and $n_{xy}$,  and the MSE for the estimator $\bar{\boldsymbol{\theta}}_{X|y}$ of the vector $\boldsymbol{\theta}_{X|y}$ as
$\text{MSE}\left(\bar{\boldsymbol{\theta}}_{X|y} \right)=\sum_{x \in \mathcal{X}}\text{MSE}\left(\bar{\theta}_{x|y} \right).$

Notice that the generative model \eqref{eq:hier_generative} is the traditional, thus non-hierarchical, Multinomial-Dirichlet model, which implies parameter independence. 
 Hence, the traditional Bayesian estimator satisfies exactly the assumptions of this model.
The Bayesian estimator is usually adopted by assuming $\tilde{\boldsymbol{\alpha}}$ to be fixed to the values of a uniform distribution on $\mathcal{X}$, i.e.,  $\boldsymbol{\alpha}^{\text{B}}=1/r\cdot \bm{1}_{\scriptscriptstyle 1 \times r}$, see e.g.  \cite{gelman2013bayesian}. However,  since in general $\tilde{\boldsymbol{\alpha}} \neq 1/r \cdot \bm{1}_{\scriptscriptstyle 1 \times r}$, the traditional Bayesian approach generates biased estimates in small samples. On the contrary, the novel hierarchical approach estimates the unknown parameter vector $\tilde{\boldsymbol{\alpha}}$ basing on 
 its posterior distribution. For this reason the proposed approach can provide estimates that are closer to the true underlying  parameters, with a particular advantage in small samples, with respect to other traditional approaches.

In order to study the MSE of different estimators, we first consider an \emph{ideal} shrinkage estimator 
\begin{equation}
\boldsymbol{\theta}_{X|y}^{*}=\tilde{\omega}_y \boldsymbol{\theta}_{X|y}^{\text{ML}}+(1-\tilde{\omega}_y)\tilde{\boldsymbol{\alpha}},
\label{eq:shrink}
\end{equation}
where $\tilde{\omega}_y \in (0,1)$ and $\theta_{x|y}^{\text{ML}}=\frac{n_{xy}}{n_y}$ is the maximum-likelihood (ML) estimator, obtained estimating  {from the observations
each vector $\boldsymbol{\theta}_{X|y}$ independently of other vectors.} This convex combination  shrinks the ML estimator towards the true underlying parameter $\tilde{\boldsymbol{\alpha}}$.
Setting $\tilde{\omega}_y=\frac{n_y}{n_y+s}$, the ideal estimator corresponds to a Bayesian estimator with known parameter $\tilde{\boldsymbol{\alpha}}$, i.e., ${\theta}^{*}_{x|y}=\frac{n_{xy}+\tilde{\alpha}_x}{n_y+s}$. 
However, since $\tilde{\boldsymbol{\alpha}}$ represents the true underlying parameter that is usually unknown, the ideal   estimator \eqref{eq:shrink} is unfeasible. 
Yet it is useful as it allows to study the MSE.

The main result concerns the comparison, in terms of MSE, of the ideal   estimator with respect to the traditional (non hierarchical) Bayesian estimator, which estimates $\tilde{\alpha}_x$ by means of the uniform distribution $1/r$, i.e., 
${\theta}^{\text{B}}_{x|y}=\frac{n_{xy}+s/r}{n_y+s}$, see e.g. \cite{gelman2013bayesian}. The traditional Bayesian estimator can be written as
\begin{equation}
\boldsymbol{\theta}_{X|y}^{\text{B}}=\omega_y \boldsymbol{\theta}_{X|y}^{\text{ML}}+(1-\omega_y)\frac{1}{r},
\label{eq:bayes}
\end{equation}
where $\omega_y=\frac{n_y}{n_y+s}$. 
\begin{thm}
Under the assumption that the true generative model is \eqref{eq:hier_generative}, the MSE for the ideal   estimator is
\begin{equation*}
\text{MSE}\left(\theta_{x|y}^{*} \right)=\bigg(\tilde{\omega}_y^2 \frac{ s }{n_y} + (1-\tilde{\omega}_y)^2\bigg) \frac{ \tilde{\alpha}_x-\tilde{\alpha}_x^2}{s+1},
\label{eq:MSE_star}
\end{equation*}
while the MSE for the traditional Bayesian estimator is
\begin{equation*}
\text{MSE}\left(\theta_{x|y}^{\text{B}} \right)=\text{MSE}\left(\theta_{x|y}^{*} \right)+(1-\omega_y^2)\left(\tilde{\alpha}_x-\frac{1}{r}\right)^2 .
\label{eq:MSE_B}
\end{equation*}
If $\tilde{\omega}_y=\omega_y$,  {$\text{MSE}\left(\boldsymbol{\theta}_{X|y}^{*} \right)\leq \text{MSE}\left(\boldsymbol{\theta}_{X|y}^{\text{B}} \right)$}.
\label{prop:MSE_sh}
\end{thm}
The proof is reported in Appendix.

Since in general $\tilde{\alpha}_x\neq\frac{1}{r}$, the second term in \eqref{eq:MSE_B} is positive and the ideal estimator achieves smaller MSE with respect to traditional Bayesian estimator.  To improve the estimates of the traditional Bayesian model in terms of MSE, we propose to act exactly on the second term of \eqref{eq:MSE_B}. Specifically, we can achieve this purpose by estimating the parameter vector $\tilde{\boldsymbol{\alpha}}$ from data, instead of considering it fixed.

The proposed hierarchical estimator defined in \eqref{eq:mean} has the same structure of the ideal   estimator \eqref{eq:shrink}:
\begin{equation*}
\hat{\boldsymbol{\theta}}_{X|y}=\omega_y \boldsymbol{\theta}_{X|y}^{\text{ML}}+(1-\omega_y)\hat{\boldsymbol{\alpha}},
\end{equation*}
where $\omega_y=\frac{n_y}{n_y+s}$. This convex combination of $\boldsymbol{\theta}_{X|y}^{\text{ML}}$ and $\hat{\boldsymbol{\alpha}}$ shrinks the ML estimator towards  the posterior  average of $\boldsymbol{\alpha}$, with a strength that is inversely proportional to $n_y$.

Contrary to the traditional Bayesian estimator \eqref{eq:bayes}, the hierarchical estimator provides an estimate of $\tilde{\boldsymbol{\alpha}}$ that converges to the true underlying parameter as $n$ increases. Indeed, 
 it is well known that the posterior  {average} $\E^D\left[\boldsymbol{\alpha}\right]$ converges to the true underlying parameter $\tilde{\boldsymbol{\alpha}}$ as $n$ goes to infinity, i.e.,  $\hat{\boldsymbol{\alpha}}\rightarrow \tilde{\boldsymbol{\alpha}}$ as $n\rightarrow +\infty$, \cite{gelman2013bayesian}. 
 As a consequence, $\hat\theta_{x|y}$ converges to $\theta_{x|y}^{*}$ and the MSE of $\hat\theta_{x|y}$ converges to the MSE of $\theta_{x|y}^{*}$, i.e., $\text{MSE}\left(\hat{\theta}_{x|y}\right)\rightarrow \text{MSE}\big({\theta}^{*}_{x|y}\big)$ as $n\rightarrow +\infty$.

In the finite sample assumption $\text{MSE}\left(\hat{\theta}_{x|y}\right)$ differs from $\text{MSE}\big({\theta}^{*}_{x|y}\big)$, since $\hat\theta_{x|y}$ includes 
an estimator of $\boldsymbol{\alpha}$. 
Since in the hierarchical model we cannot compute this quantity analytically, 
we verify by simulation that the hierarchical estimator provides good performances in terms of MSE with respect to the traditional Bayesian estimators. 

In conclusion, as we will show in the numerical experiments, the hierarchical model can achieve a smaller MSE than 
the traditional Bayesian estimators, in spite of the unfavourable conditions of the generative model \eqref{eq:hier_generative}. This gain is obtained thanks to the estimation of the parameter $\tilde{\boldsymbol{\alpha}}$, rather than considering it fixed as in the traditional approaches. In more general conditions with respect to \eqref{eq:hier_generative}, the true generating model could not satisfy parameter independence and the MSE gain of the hierarchical approach would further increase.

\section{EXPERIMENTS}
\label{sec:exp}
 In the experiments we compute the proposed hierarchical estimator
	using variational Bayes inference in R 
	by means of the \textit{rstan} package \cite{stan2}. 
The variational Bayes estimates are practically equivalent 
to those yielded by Markov Chain Monte Carlo (MCMC), though being the variational inference much more
efficient than MCMC (less than a second for estimating a CPT compared to a couple of minutes for the MCMC method).
In the following we report the results obtained via variational inference. The code is available at \url{http://ipg.idsia.ch/software.php?id=139}.
\begin{figure*}[hbt]
\includegraphics[width=0.26\textwidth]{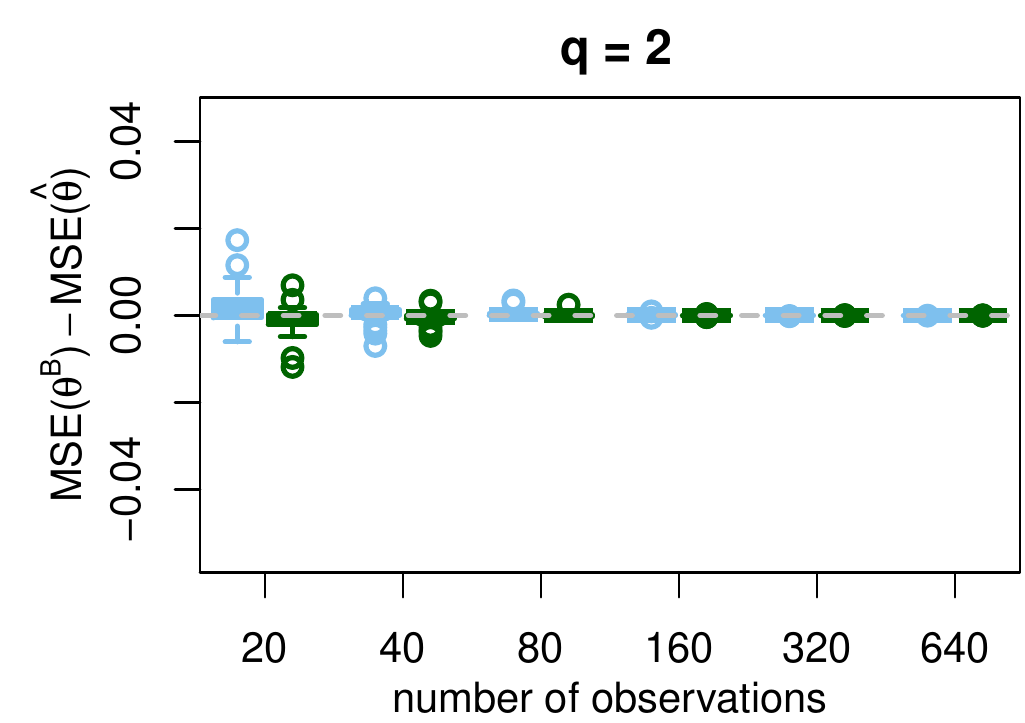}
\includegraphics[width=0.235\textwidth]{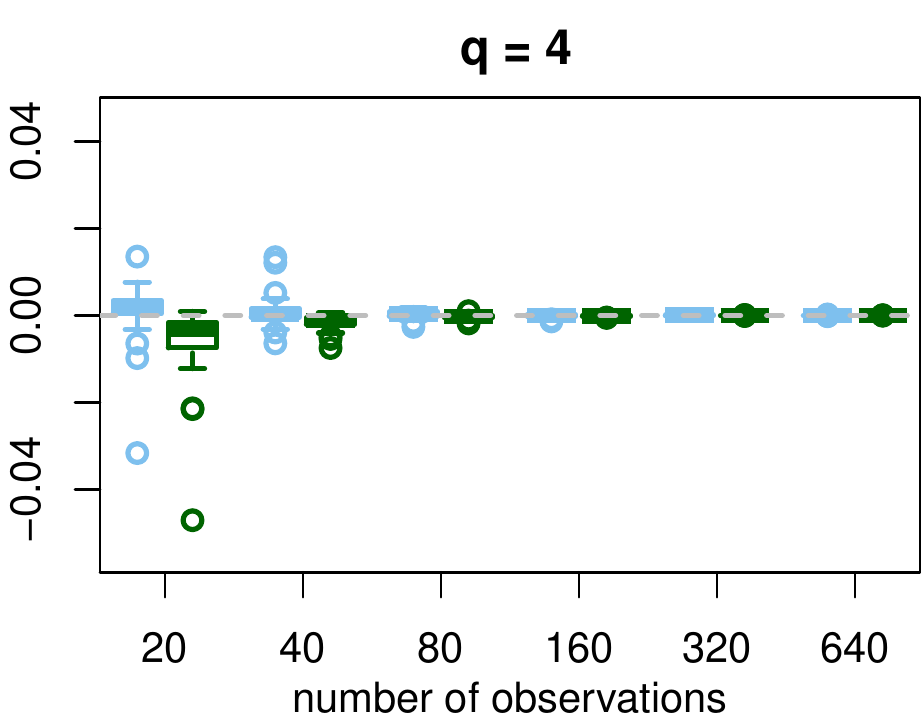}
\includegraphics[width=0.235\textwidth]{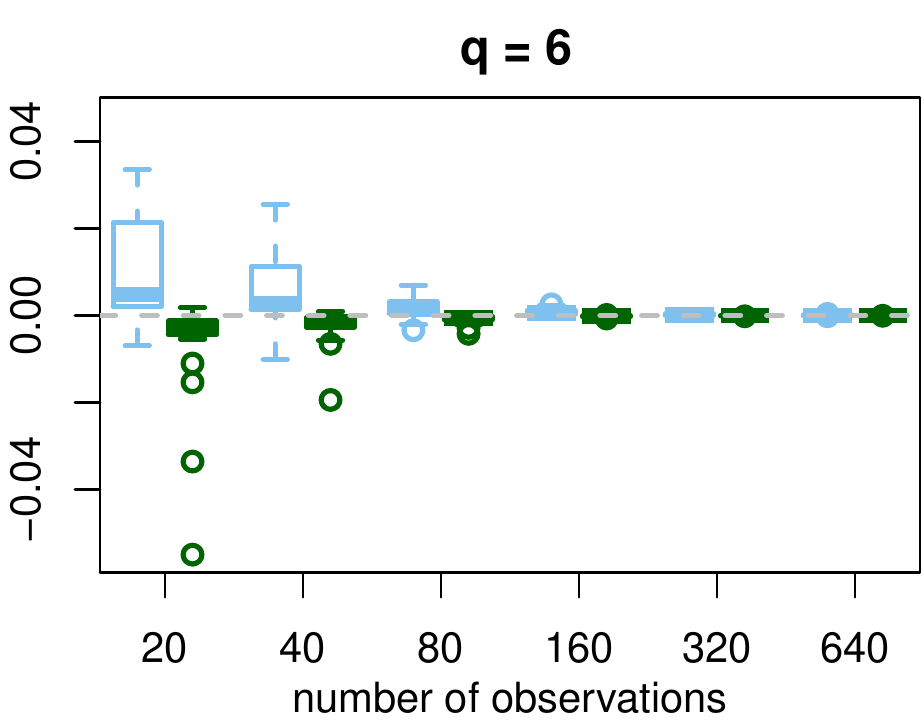}
\includegraphics[width=0.235\textwidth]{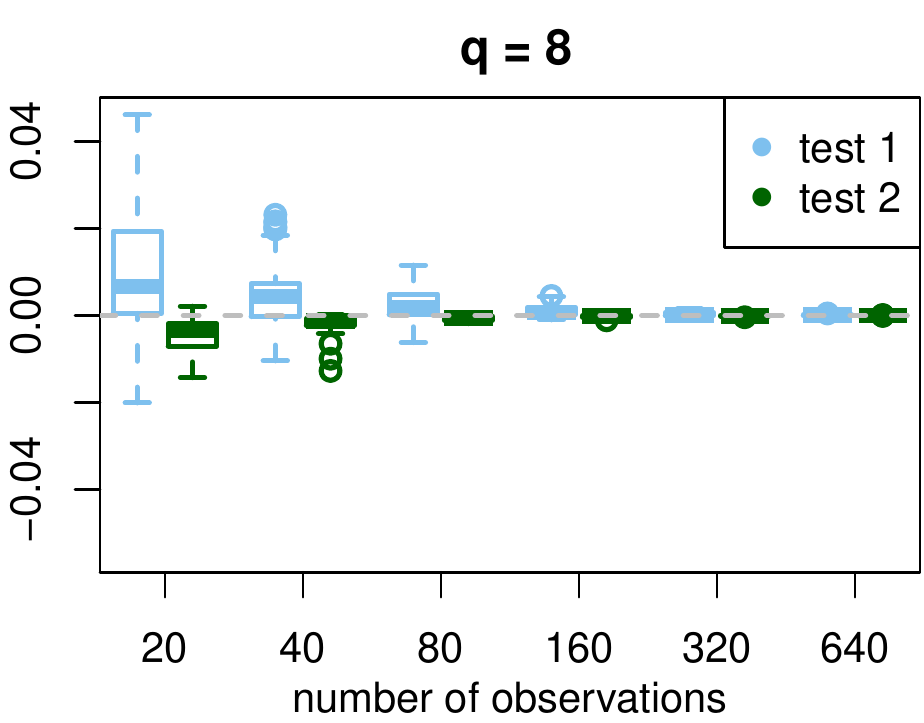}
\caption{Boxplots of MSE difference 
between the Bayesian ($s=r$) and the hierarchical estimator in test 1 (light blue) and test 2 (green) with different dimension of the conditioning set ($q=2, 4, 6, 8$). 
Positive values favour the hierarchical model.}
\label{fig:MSE}
\end{figure*}

\subsection{MSE analysis}
In the first study we assess the performances of the hierarchical estimator in terms of MSE.
We consider two different  settings, in which we generate observations from model \eqref{eq:hier_generative}, where $\tilde{\boldsymbol{\alpha}}$ is fixed. In the first setting (test 1) 
we sample
$\tilde{\boldsymbol{\alpha}}$ from a Dirichlet distribution with parameter $\bm{1}_{\scriptscriptstyle 1 \times r}$, while in the second setting (test 2) we sample it from a Dirichlet distribution with parameter $10^6 \cdot\bm{1}_{\scriptscriptstyle 1 \times r}$. 
Under test 2 the parameters of the sampling distribution for $\tilde{\boldsymbol{\alpha}}$ are very large and equal to each other, implying
 $\tilde{\alpha}_x\simeq 1/r$, $\forall x \in \mathcal{X}$. For this reason, test 2 is the ideal setting 
for the traditional Bayesian estimator, while test 1 is the ideal setting for the hierarchical estimator.
 
In both test 1 and test 2 we consider all the possible combinations of $r$ (the number of states of $X$)  and $q$  (the number of conditioning states), with $r\in \left\{ 2, 4, 6, 8\right\}$ and $q\in \left\{ 2, 4, 6, 8\right\}$. For each combination of $r$ and $q$, for both test 1 and test 2,  we generate 
data sets with size
$n\in \left\{20, 40, 80, 160, 320, 640\right\}$.
We repeat the data sampling and the estimation procedure 10 times for each 
combination of $r$, $q$ and $n$.
Then we compare the estimates yielded
 by the hierarchical estimator, with $s=r$ and $\bm{\alpha}_0=\mathbf{1}_{1\times r}$, 
 and by the traditional Bayesian estimator assuming parameter independence, with $s=r$. We compare the performance of different estimators by computing the difference in terms of average MSE, defined  as
$\text{MSE}(\bar{\boldsymbol{\theta}}_{X|Y})={\sum_{y \in \mathcal{Y}}\sum_{x\in \mathcal{X}}\text{MSE}(\bar{\theta}_{x|y})}/{r q}$ for an estimator $\bar{\boldsymbol{\theta}}_{X|Y}$.
In every repetition of the experiment $\left\{1, \ldots, 10\right\}$, we then compute $\text{MSE}(\boldsymbol{\theta}^{\text{B}}_{X|Y})- \text{MSE}(\hat{\boldsymbol{\theta}}_{X|Y})$ and we represent it in Figure \ref{fig:MSE}. 

The results show that the hierarchical estimator mostly provides better or equivalent results in comparison to 
$\boldsymbol{\theta}^{\text{B}}_{X|Y}$, 
especially for small $n$ and/or large $q$. In a Bayesian network it is usual to have large values of $q$, since $q$ represents the cardinality of the parents' joint states set.
In test 1 (light blue boxplots) the advantage of the hierarchical estimator over the Bayesian one is generally large, as expected. 
The advantage of the hierarchical model steadily increases as $q$ increases,  becoming relevant for 
$q=6$ or $q=8$. 
For large $n$ the gap between the two estimators vanishes, although it is more persistent when dealing with large $q$.
Interestingly, in test 2 (green boxplots), the traditional Bayesian estimator is just slightly better than the hierarchical one, even though the former is derived exactly from the true generative model. 
The traditional estimator has a small advantage only 
for  {$q=4$} and small values of $n$, and this advantage quickly decreases if either $q$ or $n$ increase.

\subsection{Joint distribution fitting}
\label{sec:joint}
In the second study we assess the performance of the hierarchical estimator in the recovery of the joint distribution of a given Bayesian network.

We consider 5 data sets from UCI Machine Learning Repository: \emph{Adult}, \emph{Letter Recognition}, \emph{Nursery}, \emph{Pen-Based Recognition of Handwritten Digits} and \emph{Spambase}. 
We discretise all numerical variables into five equal-frequency bins
and we consider only instances without missing values.
For each dataset we first learn,  from all the available data, the  associated directed acyclic graph (DAG) by means of a hill-climbing greedy search, as 
implemented in the \textit{bnlearn} package \cite{scutari2010learning}.
We then keep such structure as fixed for all the experiments referring to the same data set,  {since our focus is not structural learning}.
Then, for each data set and for each $n\in \left\{20, 40, 80,160,\right.$ $\left. 320, 640, 1280\right\}$ we repeat 10 times the procedure of 1)
sampling $n$ observations from the data set and 2)
estimating the CPTs. 
We perform estimation using the proposed hierarchical approach, with $s=r$ and $\bm{\alpha}_0=\mathbf{1}_{1\times r}$, and the traditional
BDeu prior (Bayesian estimation under parameter independence) with $s=1$ and $s=10$.
The choice of $s=1$ is the most commonly adopted  {in practice}, while $s=10$  is the default value proposed by the \textit{bnlearn} package.  Conversely, we did not offer any choice to the hierarchical model. Indeed,  we set the smoothing factor $s$ in the proposed model to the number of states of the child variable, which has the same order of magnitude of the smoothing factors used in the traditional Bayesian approach. In spite of the more limited choice for the parameter $s$
, the hierarchical estimator consistently outperforms the traditional Bayesian estimator, regardless whether the latter adopts a smoothing factor 1 or 10.

We then measure the log-likelihood of all the instances included in the test set, where the test contains all the instances not present in training set. 
We report in the top panels of Figure \ref{fig:UCI_datasets} the difference between the log-likelihood
of the
hierarchical approach and the log-likelihood of Bayesian estimation under local parameter independence, i.e., the log-likelihood ratio.
The log-likelihood ratio approximates the 
ratio between the
Kullback-Leibler (KL) divergences of the two models. The KL of a given model measures the distance
between the estimated and the true underlying
joint distribution.

The log-likelihood ratios obtained in the experiments are extremely large on small sample sizes, being larger than  \textit{one thousand} on all data sets (Figure \ref{fig:UCI_datasets}, top panels).
This shows the huge gain delivered by the hierarchical approach when dealing with small data sets.
This happens regardless of the equivalent sample size used to perform Bayesian estimation under parameter independence:
we note however that in general $s=10$ yields better results than $s=1$.
The lowest gains are obtained on the data set \emph{Nursery};
the reason is that the DAG of  \emph{Nursery} has the lowest number of parents per node (0.9 on average, compared to about twice as much for the other DAGs). Thus, this data set  is the less challenging from the parameter estimation viewpoint, with respect to the others.
We point out that significant likelihood ratios are obtained in general even for large samples, even though they are not apparent
from the figure due to the scale.
For instance,  for  $n=320$ the log-likelihood ratios range from
50 (\emph{Nursery}) to 85000 (\emph{Letter}).

\begin{figure*}[htbp]
\includegraphics[width=0.2\textwidth]{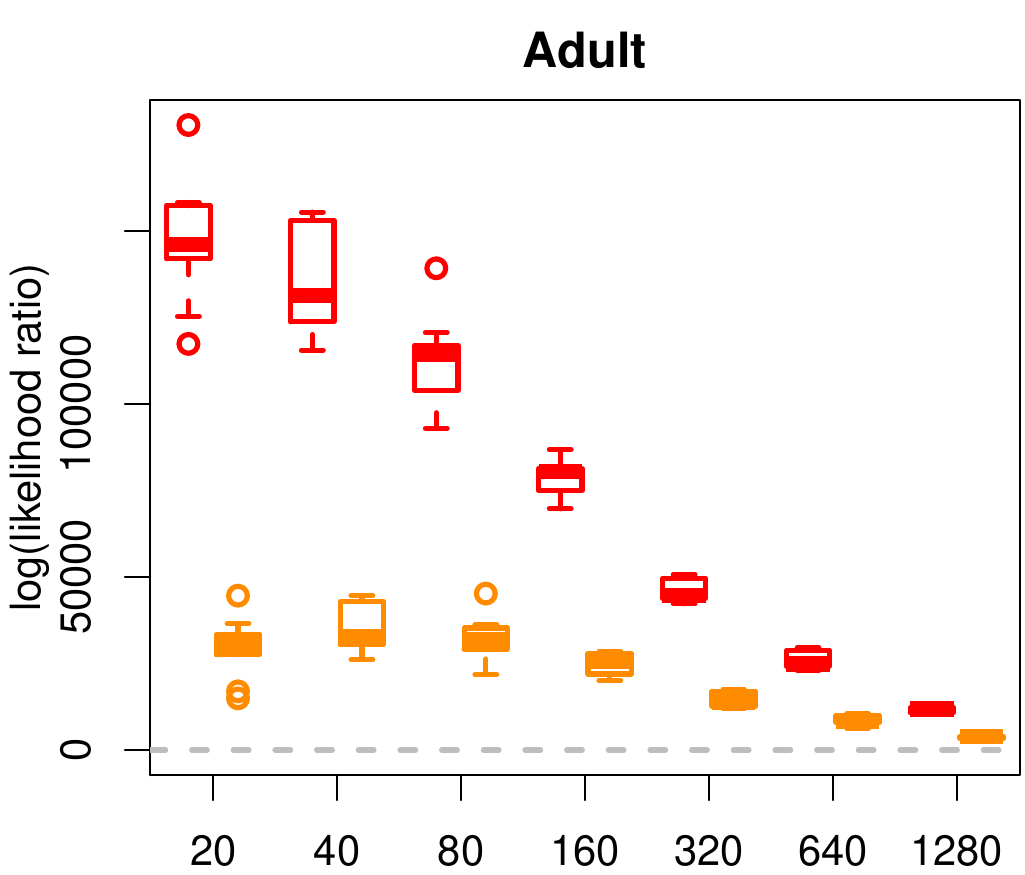}
\includegraphics[width=0.19\textwidth]{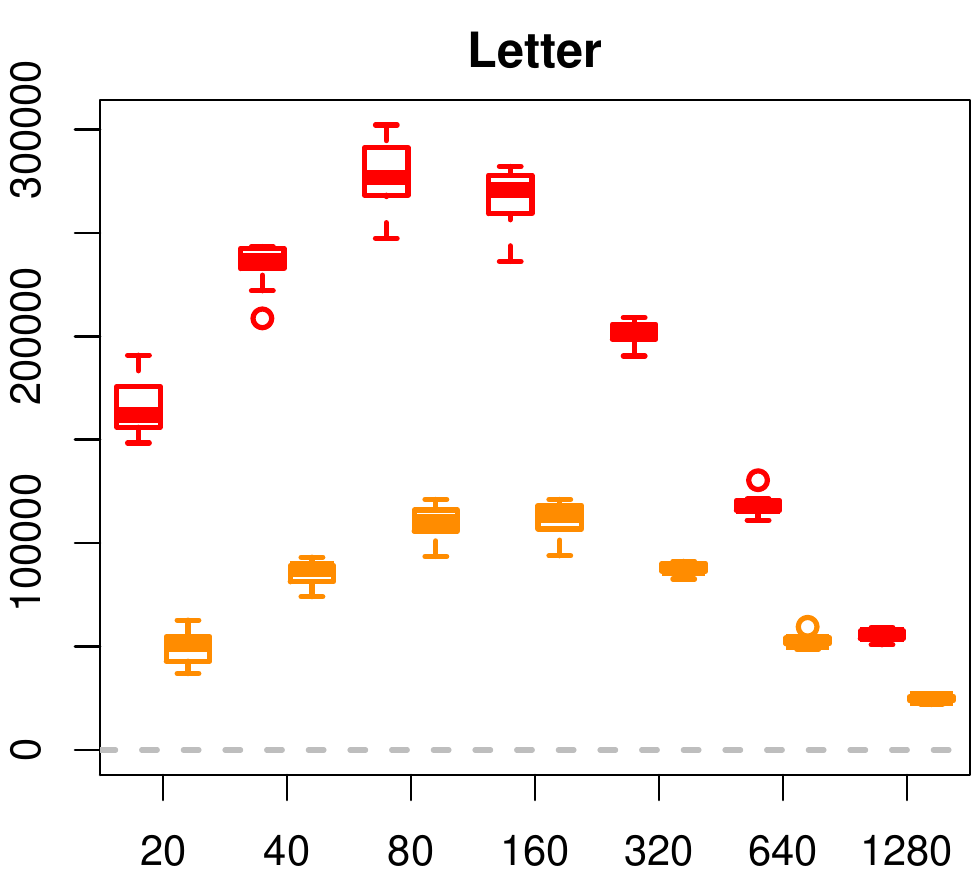}
\includegraphics[width=0.19\textwidth]{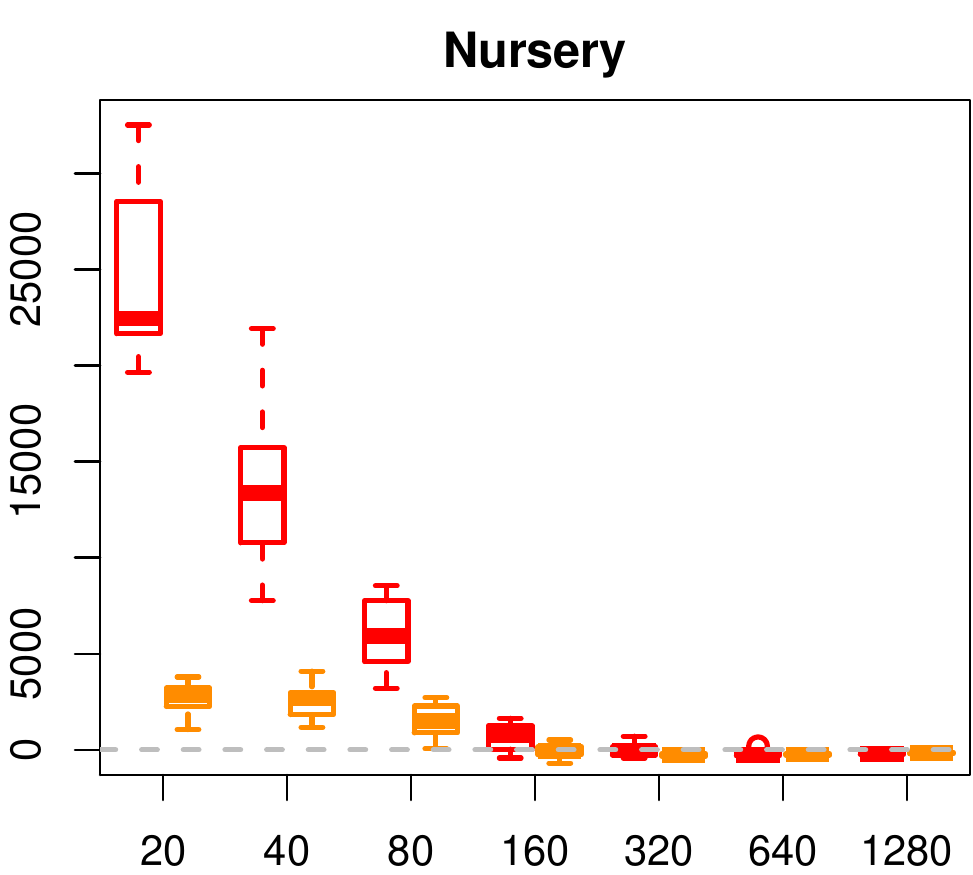}
\includegraphics[width=0.19\textwidth]{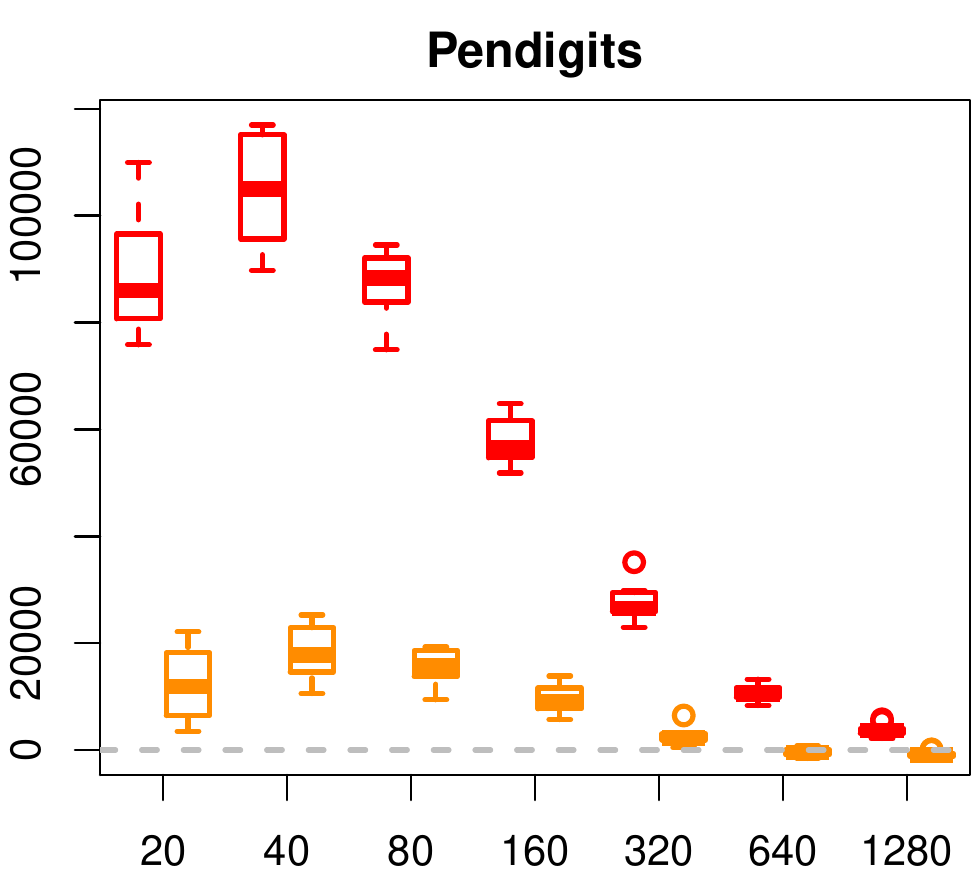}
\includegraphics[width=0.19\textwidth]{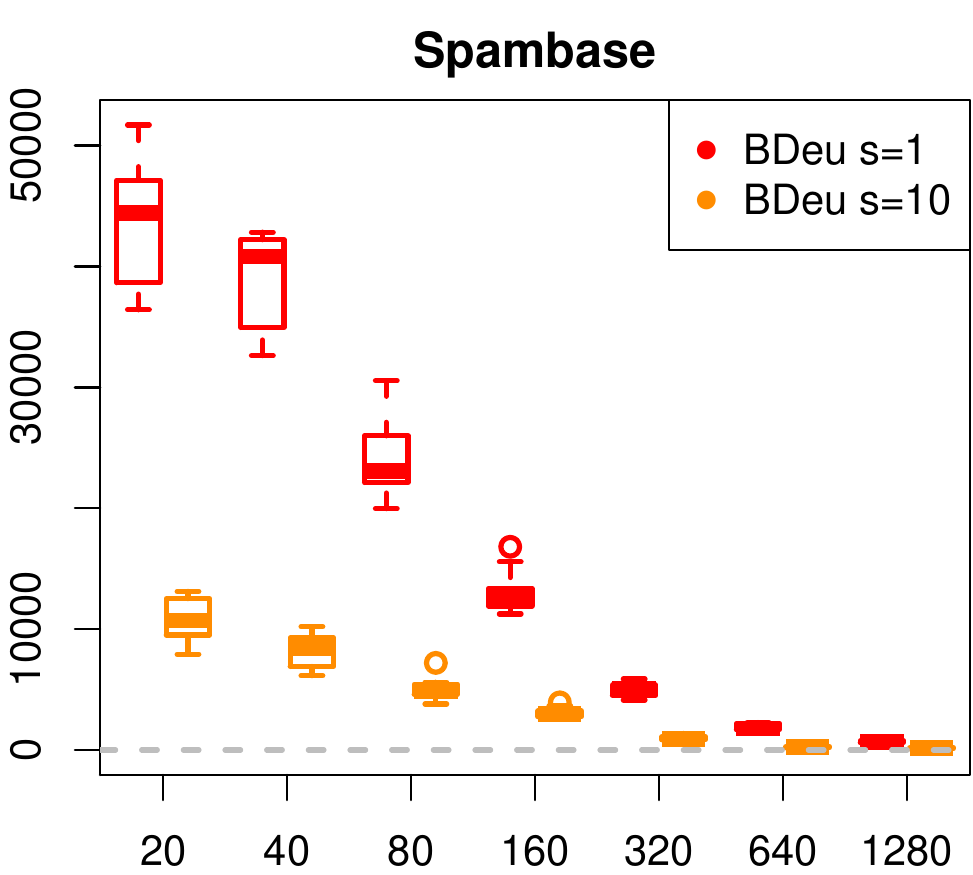}\\
\includegraphics[width=0.2\textwidth]{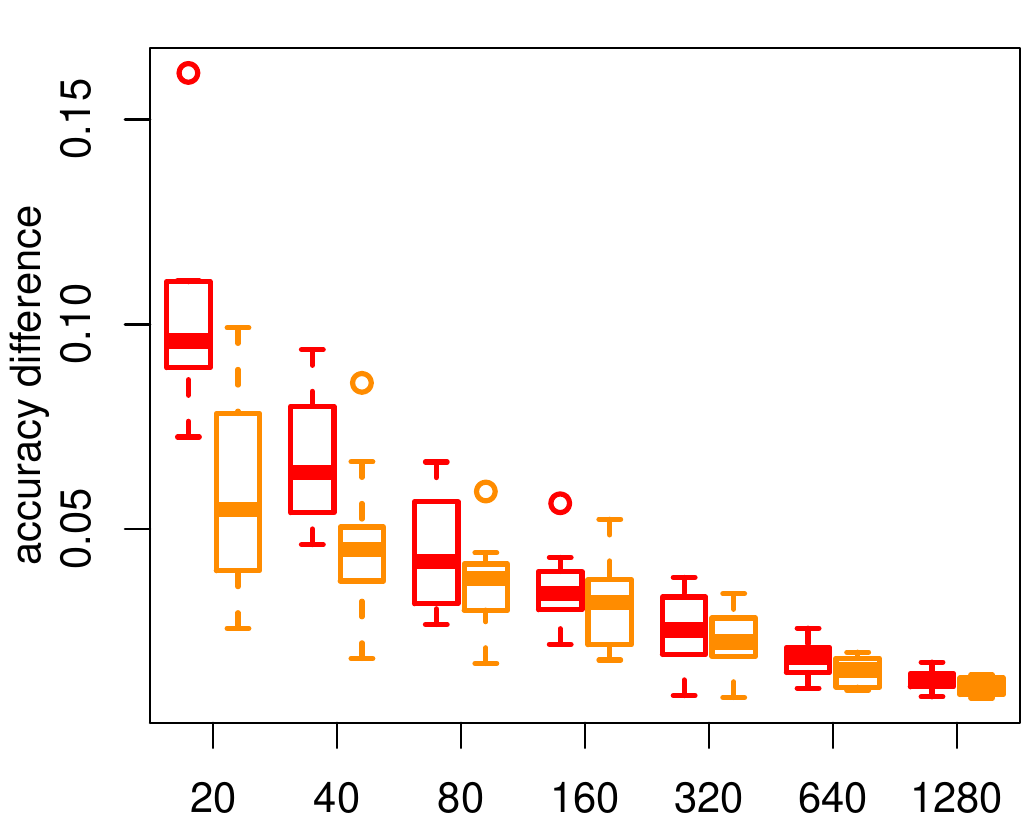}
\includegraphics[width=0.19\textwidth]{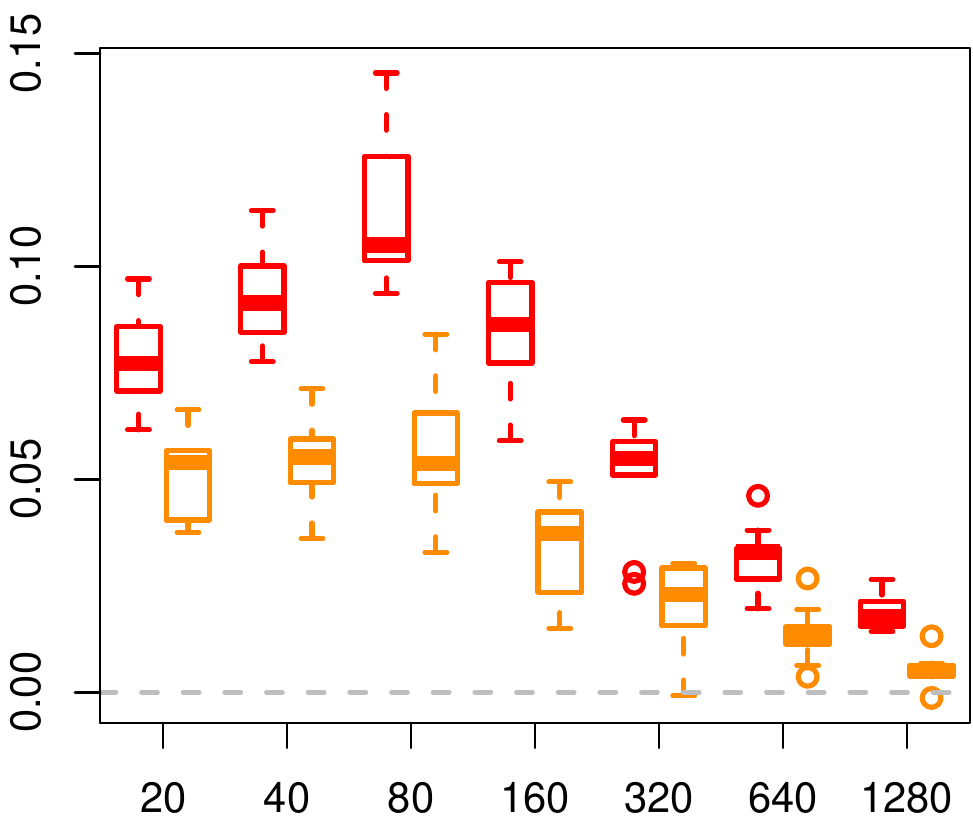}
\includegraphics[width=0.19\textwidth]{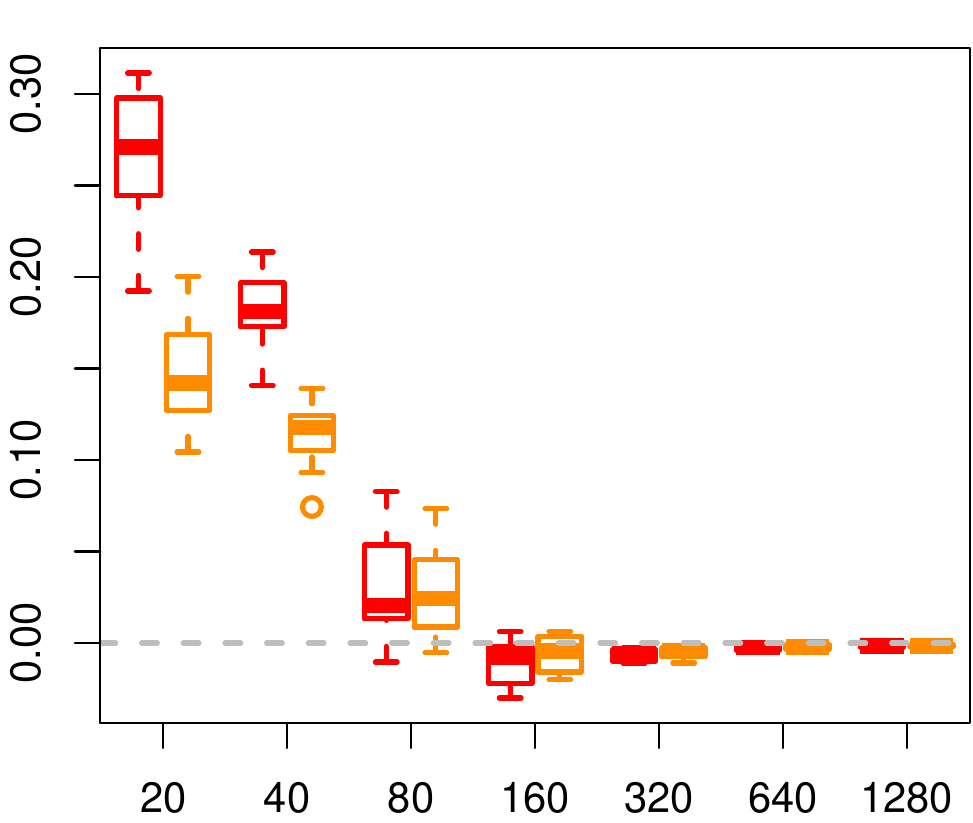}
\includegraphics[width=0.19\textwidth]{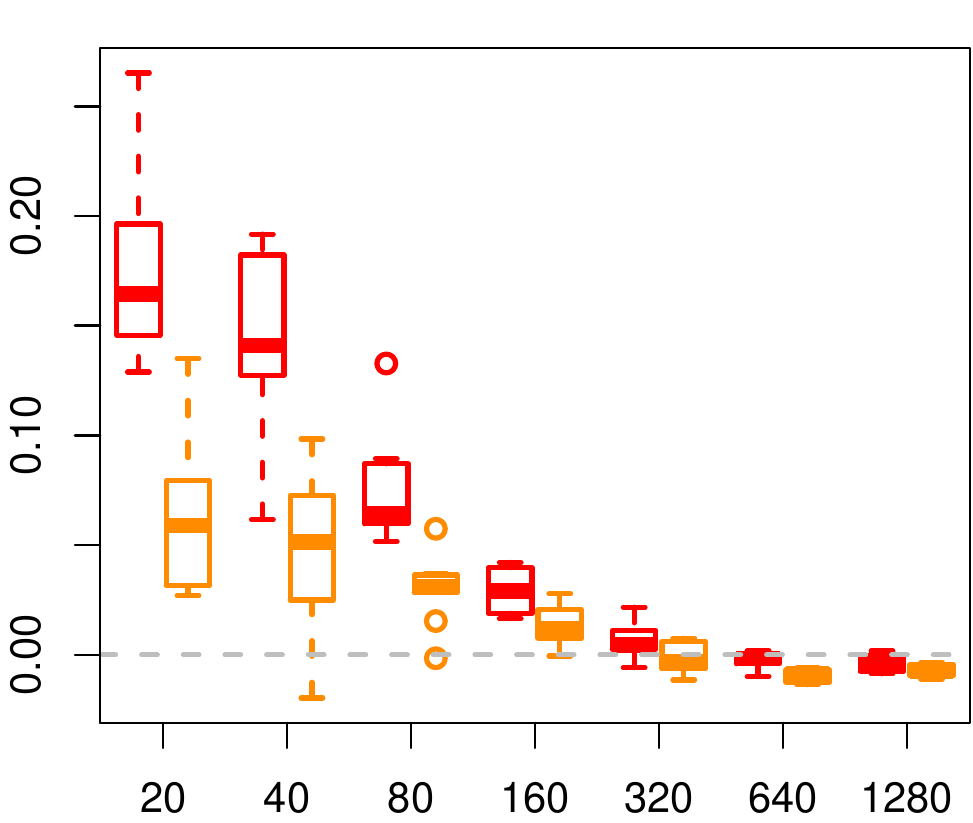}
\includegraphics[width=0.19\textwidth]{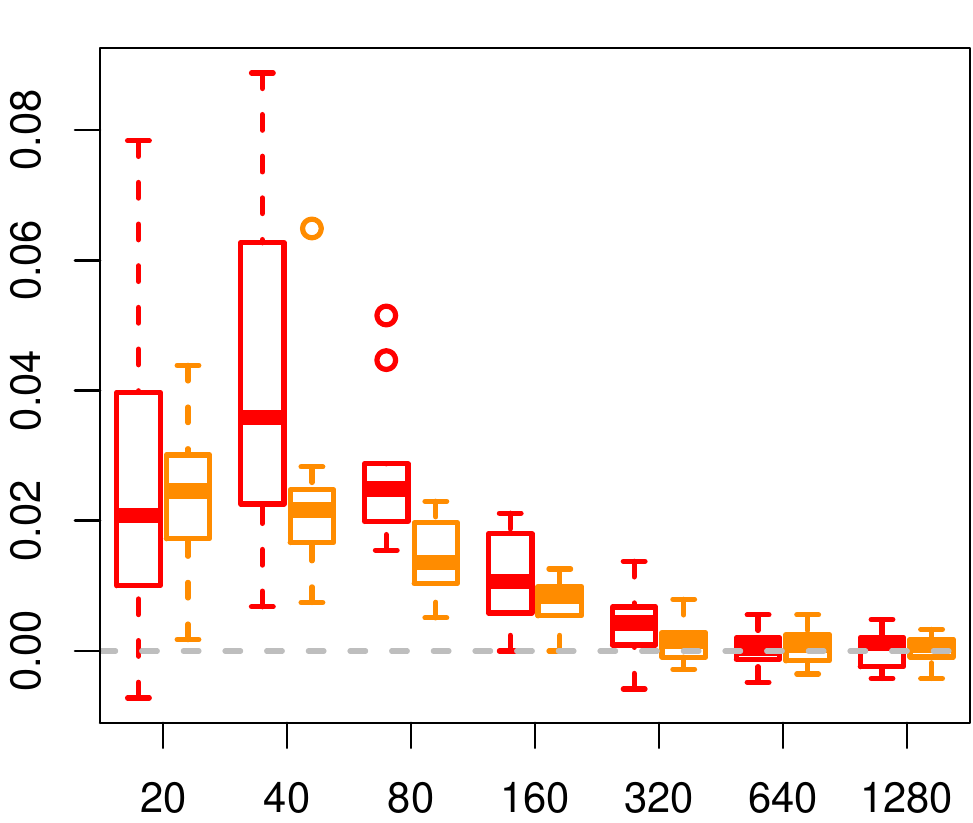}\\
\includegraphics[width=0.2\textwidth]{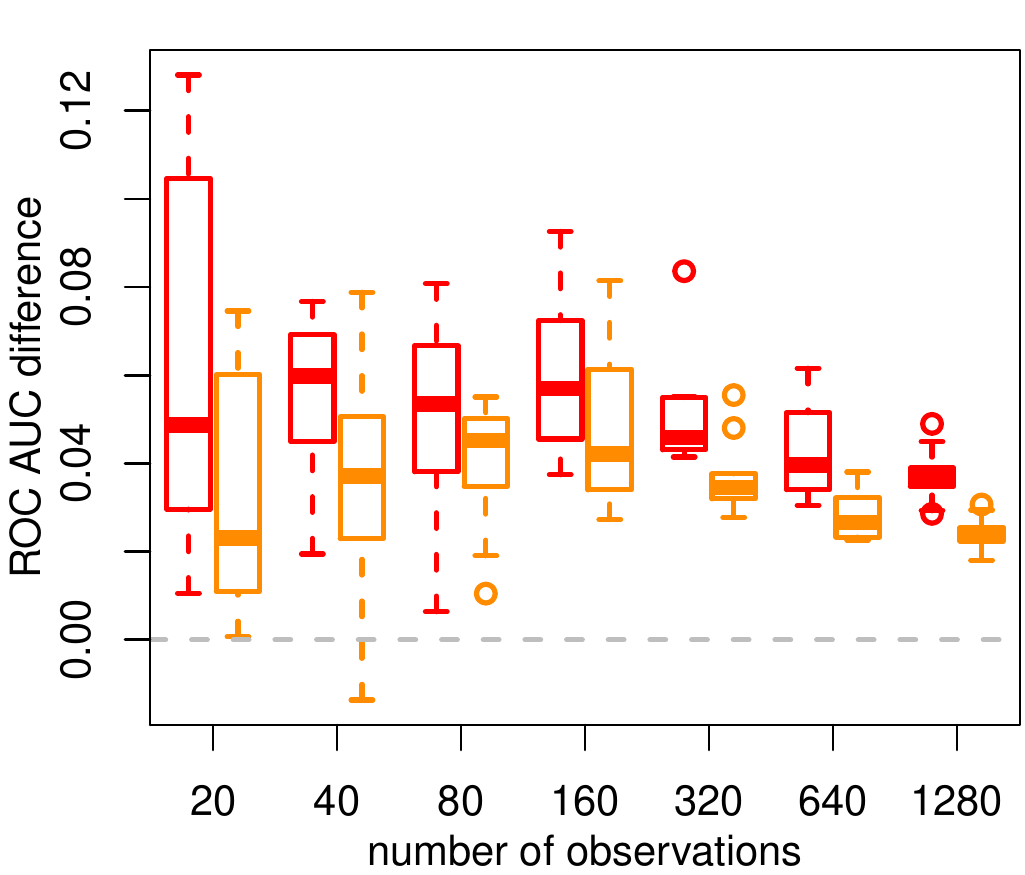}
\includegraphics[width=0.19\textwidth]{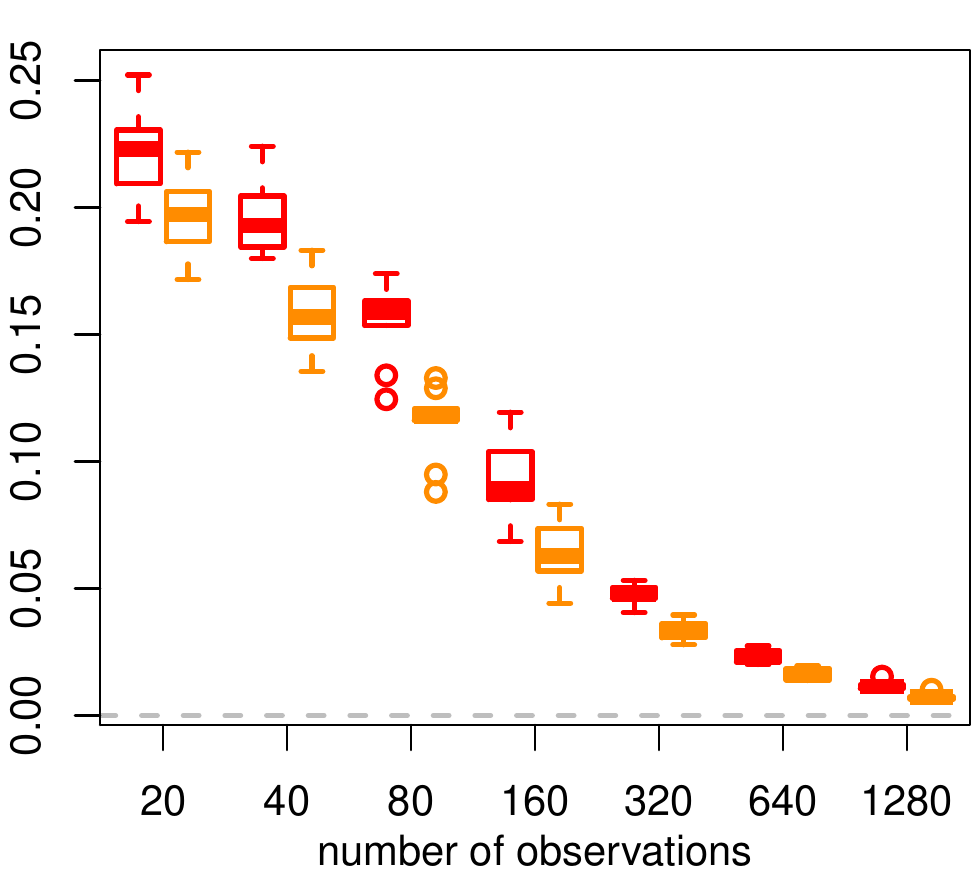}
\includegraphics[width=0.19\textwidth]{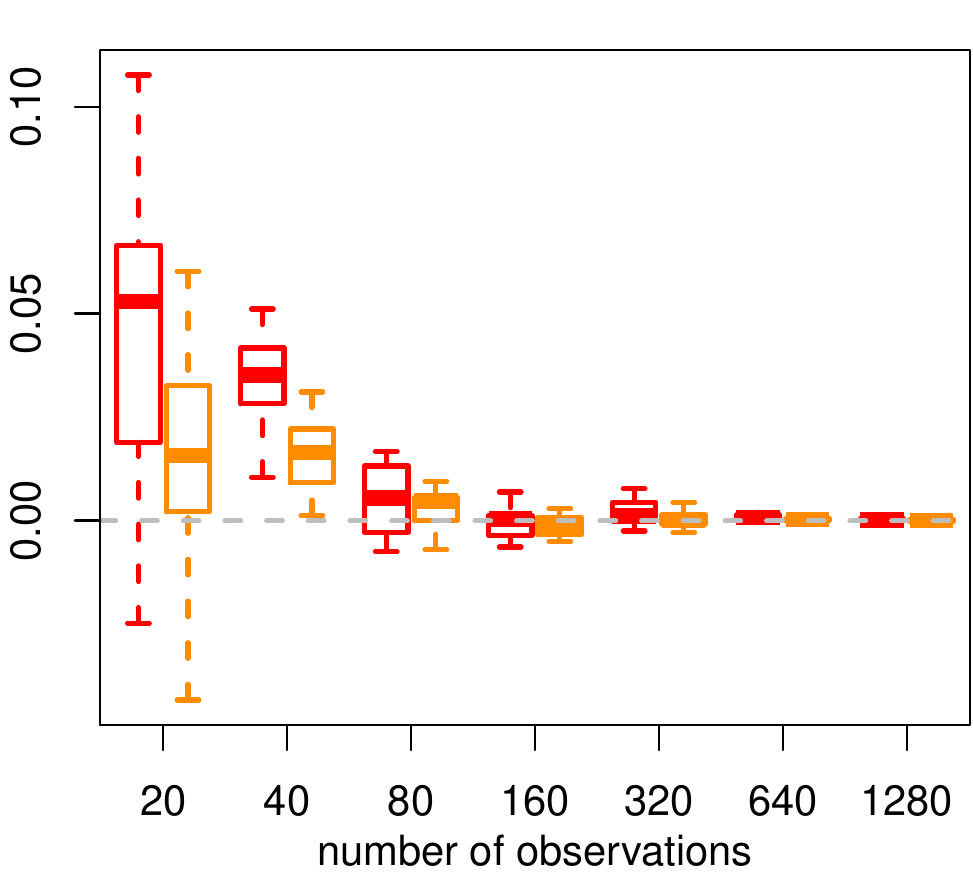}
\includegraphics[width=0.19\textwidth]{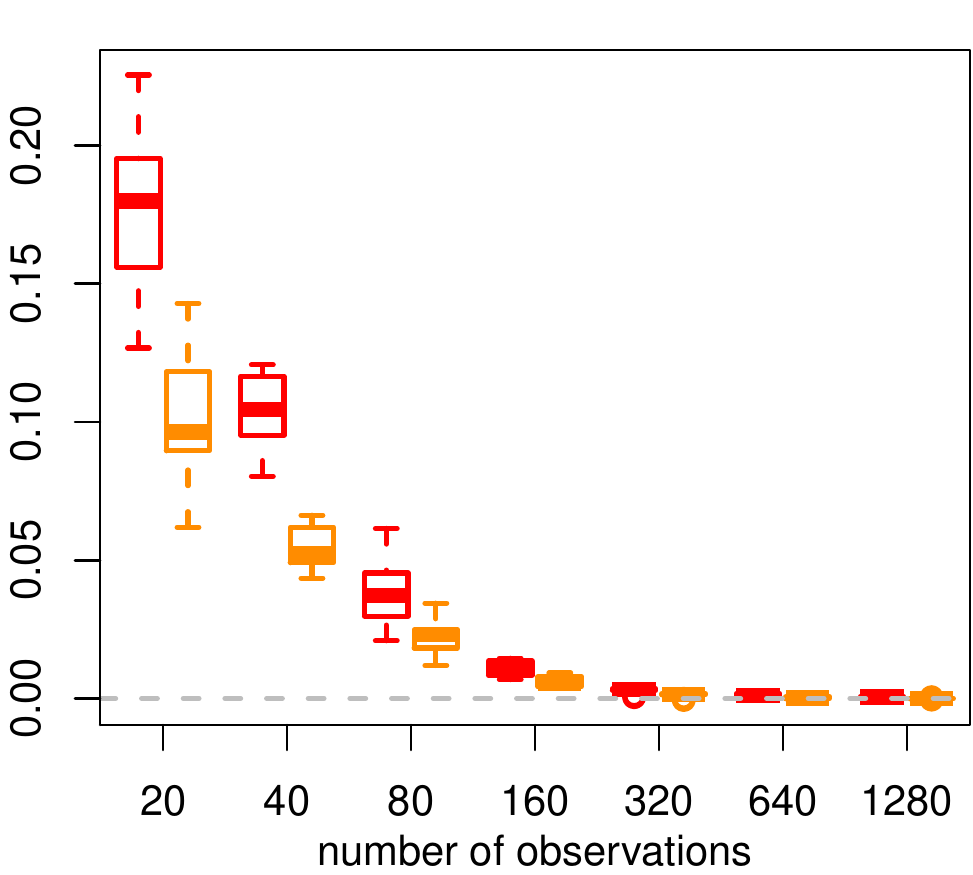}
\includegraphics[width=0.19\textwidth]{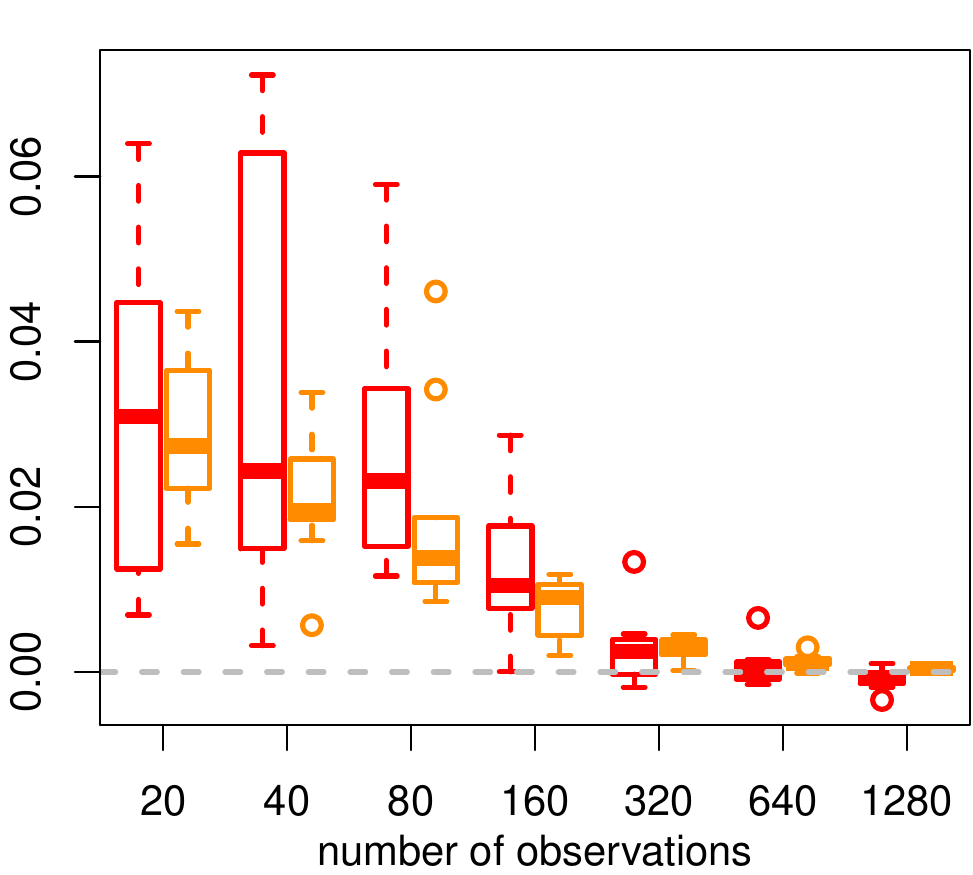}
\caption{Boxplots of the logarithm of the likelihood ratio (top panels), accuracy gain (central panels) and area under the ROC gain (bottom panels) obtained comparing the hierarchical method with respect to BDeu ($s=1$ in orange and $s=10$ in red) for the five machine learning datasets analysed. Positive values favour the hierarchical model.}\label{fig:UCI_datasets}
\end{figure*}

\subsection{Classification}
In the third study we assess the performance of the hierarchical estimator in terms of classification.
We consider the same datasets of the previous experiment, discretised in the same way. 

For each dataset we first learn the Tree-Augmented Naive Bayes (TAN) structure  by means of the \textit{bnlearn} package. The networks are estimated on the basis of all the available samples and are kept fixed for all the experiments
referring to the same data set,  since our focus is not structural learning. 
Then, for each dataset and for each $n\in \left\{20, 40, 80, 160,\right.$ $\left. 320, 640, 1280\right\}$, we sample $n$ observations. 
We then estimate  the CPTs of the Bayesian network from the sampled data by means of the hierarchical estimator ($s=r$ and $\bm{\alpha}_0=\mathbf{1}_{1\times r}$) and the traditional Bayesian estimators obtained under a  BDeu prior ($s=1$ and $s=10$). We repeat the sampling and the estimating steps 10 times for each value of $n$ and each data set. We then classify each instance of the test set, which contains 1000 instances sampled uniformly from all the instances not included in the training set.
We assess the classification performance by measuring accuracy and  {area under the ROC (ROC AUC)} of the classifiers. 
In the central panels of Figure \ref{fig:UCI_datasets} we report the difference in accuracy between the hierarchical estimator and the traditional Bayesian ones, while in the bottom panels of the same figure we report the difference in ROC AUC between the same classifiers. 

The area under the ROC is a more sensitive indicator for the correctness of the estimated posterior probabilities with respect to accuracy. 
According to Figure \ref{fig:UCI_datasets} (bottom panels),
the hierarchical approach yields consistently higher ROC AUC than both the BDeu classifiers.
The increase of ROC AUC in small samples ($n=20$, $n=40$) ranges between  \textit{2} and \textit{20} points compared to both the BDeu priors. As $n$ increases this gain in ROC AUC tends to vanish. However, for $n>320$ the gain in ROC AUC for the datasets \emph{Adult} and \emph{Letter} ranges between 1 and 5 points.

Figure \ref{fig:UCI_datasets} (central panels) shows also improvements in accuracy, even if this indicator is less sensitive to the estimated posterior probability than the area under ROC. Indeed, in computing the accuracy, the most probable class is compared to the actual class,
without paying further attention to its posterior probability. In small samples ($n=20$, $n=40$) there is an average increase of accuracy of about \textit{5} points compared to the BDeu prior with $s=10$ and of about \textit{10} points 
compared to the BDeu prior with $s=1$. The accuracy improvements tends to decrease as $n$ increases; 
yet on both \emph{Adult} and \emph{Letter} data sets an accuracy improvement of about 1-2 points is shown also for $n=1280$.

\section{CONCLUSIONS} 
\label{sec:concl}
We have presented a novel approach for estimating the conditional probability tables by relaxing the local independence assumption.
Given the same network structure,  the novel approach  yields a consistently better fit to the joint distribution than the traditional Bayesian estimation under parameter independence; it also improves classification performance.
Moreover, the introduction of variational inference makes the proposed method competitive in terms of computational cost  with respect to the traditional Bayesian estimation.

\appendix

In order to prove Theorem \ref{thm:theta_mean}, we first need to derive some results concerning the posterior moments for the vector $\boldsymbol{\alpha}$, whose general element $\alpha_x$ is associated to state $x \in \mathcal{X}$ for the variable $X$.  {Given a dataset $D$,} the $k$-th posterior moment for the element $\alpha_x$ is 
\begin{equation*}
\E\left[\left. \alpha^k_{x}\right| D\right]=
\int{ \alpha^k_x p(\left.\boldsymbol{\alpha}\right | D) d \boldsymbol{\alpha}}.
\end{equation*}
The following proposition states a general result for computing any posterior moment of $\boldsymbol{\alpha}$, whose general expression is
$\E\left[\prod_{x \in \mathcal{X}} \alpha^{k_x}_{x}| D\right]$,
where $k_x \in \mathbb{N}$ represents the power of element $\alpha_x$. 

\begin{lem}
Under the assumptions of model (2), the posterior  {average} of the quantity $\prod_{x \in \mathcal{X}} \alpha^{k_x}_{x}$, with $k_x \in \mathbb{N}$ $\forall x \in \mathcal{X}$, is
\begin{equation}
\E^{D}\left[ \prod_{x \in \mathcal{X}} \alpha^{k_x}_{x}\!\right]\!=\!\gamma \!\!
\int{\!\!\! \prod_{x \in \mathcal{X}}\!\!\!\! \prod_{\substack{y \in \mathcal{Y}\\ \text{s.t. } n_{xy}>0}} 
\!\!\!\! \prod_{\nu=1}^{n_{xy}} \!\!\left(\!s \alpha_x+\nu-1\!\right) \alpha_x^{\tilde{k}_x}
d \boldsymbol{\alpha}},
\label{eq:anymoment}
\end{equation}
where $\tilde{k}_x=\alpha_{0,x}+k_x-1$ and 
$\gamma$ is a proportionality constant such that
\begin{equation}
\gamma^{-1}= \int{ \prod_{x \in \mathcal{X}} \prod_{\substack{y \in \mathcal{Y}\\ \text{s.t. } n_{xy}>0}} 
\prod_{\nu=1}^{n_{xy}} \left(s \alpha_x+\nu-1\right)  \alpha_x^{\alpha_{0,x}-1}
d \boldsymbol{\alpha}}.
\label{eq:gamma}
\end{equation}
\label{thm:moments}
The element $x'$ of the posterior  {average} vector $\E^{D}\left[\boldsymbol{\alpha}\right]$ is
\begin{align}
\hat{\alpha}_{x'}
\!=\!  \gamma \!\!
\int{\!\! \prod_{x \in \mathcal{X}} \!\!\!\! \prod_{\substack{y \in \mathcal{Y}\\ \text{s.t. } n_{xy}>0}} 
\!\!\!\! \prod_{\nu=1}^{n_{xy}} \left(s \alpha_x+\nu-1\right) \alpha_x^{\delta_{x, x'}}
\!d \boldsymbol{\alpha}},
\label{eq:alpha}
\end{align}
where 
$\delta_{x, x'}$ is a Kronecker delta.\\
The element $(x', x'')$ of the posterior covariance matrix $\Cov^{D}{\left(\boldsymbol{\alpha} \right)}$ is
\begin{equation}
\Cov^{D}{\left(\alpha_{x'}, \alpha_{x''}\right)}=
\E^{D}\left[\alpha_{x'}\alpha_{x''}\right]-
\hat{\alpha}_{x'}\hat{\alpha}_{x''},
\label{eq:cov_alpha}
\end{equation}
where
\begin{equation*}
\E^{D}\!\left[\!\alpha_{x'}\alpha_{x''}\!\right]\!= \! \gamma \!\!
\int{\!\!\! \prod_{x \in \mathcal{X}}\!\!\!\!\! \prod_{\substack{y \in \mathcal{Y}\\ \text{s.t. } n_{xy}>0}} 
\!\!\!\!\!\prod_{\nu=1}^{n_{xy}} \left(s \alpha_x+\nu-1\right) \alpha_x^{\delta_{x, x'}+\delta_{x, x''}}
\!\!d \boldsymbol{\alpha}}.
\end{equation*}
\end{lem}
Both integrals in \eqref{eq:anymoment} and  \eqref{eq:gamma} are multiple integrals computed with respect to the $r$ elements of vector $\bm{\alpha}$, such that $\sum_{x \in \mathcal{X}}\alpha_x=1$.  {The space of integration is thus the standard $r$-simplex.} 

\begin{proof}[Proof of Lemma \ref{thm:moments}]
Under the assumptions of model (2), the joint posterior density of $\boldsymbol{\alpha}, \boldsymbol{\theta}_{X|y_1}, \ldots, \boldsymbol{\theta}_{X|y_q} $ is
\begin{align*}
& p(\boldsymbol{\alpha}, \boldsymbol{\theta}_{X|y_1}, \ldots, \boldsymbol{\theta}_{X|y_q} | 
D
)   \\
&  
 \propto \frac{\Gamma\left(s\right)}{\prod_{x \in \mathcal{X}}\Gamma\left(s\alpha_x\right)} \prod_{y \in \mathcal{Y}} \prod_{x \in \mathcal{X}} (\theta_{x|y})^{n_{xy}+s\alpha_x-1}\alpha_x^{\alpha_{0,x}-1}.
\end{align*}
Marginalising $p(\boldsymbol{\alpha}, \boldsymbol{\theta}_{X|y_1}, \ldots, \boldsymbol{\theta}_{X|y_q} | 
D)$ with respect to $\boldsymbol{\theta}_{X|y_1}, \ldots, \boldsymbol{\theta}_{X|y_q}$, we obtain the marginal posterior density for $\boldsymbol{\alpha}$, i.e.,
\begin{align}
& p(\boldsymbol{\alpha} |  
D)\!\!  \propto \!\! 
\frac{\Gamma\left( s\right)}{\prod_{x \in \mathcal{X}}\! \Gamma\left(s \alpha_x\right)}\!\! \prod_{y \in \mathcal{Y}}\! \!  \frac{\prod_{x \in \mathcal{X}} \!\Gamma\left(s \alpha_x\!\!+\!n_{xy}\right)}{\Gamma\left( s+  n_y\right)}\!\!\prod_{x \in \mathcal{X}} \! \!\alpha_x^{\alpha_{0,x}-1}.
\label{eq:marginal_alpha}
\end{align}
Thanks to the well-known property of the Gamma function
\begin{equation*}
\Gamma(\alpha+m)=\prod_{\nu=1}^m(\alpha+\nu-1)\cdot \Gamma(\alpha), \ \ \text{for }m\geq 1
\label{eq:Gamma}
\end{equation*}
we can write the posterior marginal density \eqref{eq:marginal_alpha} as
\begin{align}
p(\boldsymbol{\alpha} |  
D
) 
\propto \prod_{x \in \mathcal{X}}  \prod_{\substack{y \in \mathcal{Y}\\ \text{s.t. } n_{xy}>0}} \prod_{\nu=1}^{n_{xy}} \left(s \alpha_x+\nu-1\right)
 \alpha_x^{\alpha_{0,x}-1}.
 \label{eq:marginal_alpha_proof}
\end{align}
The proportionality constant of the posterior marginal density is obtained by integrating the right term in \eqref{eq:marginal_alpha_proof} with respect to the $r$ elements of $\boldsymbol{\alpha}$, such that $\sum_{x \in \mathcal{X}}\alpha_x=1$. The resulting proportionality constant is thus
\begin{equation*}
\gamma= \left(\int{ \prod_{x \in \mathcal{X}} \prod_{\substack{y \in \mathcal{Y}\\ \text{s.t. } n_{xy}>0}} \prod_{\nu=1}^{n_{xy}} \left(s \alpha_x+\nu-1\right) \alpha_x^{\alpha_{0,x}-1}
d \boldsymbol{\alpha}}\right)^{-1}.
\end{equation*}

The posterior  average for the quantity $\prod_{x \in \mathcal{X}} \alpha^{k_x}_{x}$ can be derived directly from the posterior marginal density of $\boldsymbol{\alpha}$ as
\begin{equation*}
\E^{D}\left[ \prod_{x \in \mathcal{X}} \alpha^{k_x}_{x}\right]=\gamma \!\!
\int{\!\! \prod_{x \in \mathcal{X}} \prod_{\substack{y \in \mathcal{Y}\\ \text{s.t. } n_{xy}>0}} \prod_{\nu=1}^{n_{xy}} \left(s \alpha_x+\nu-1\right) \alpha_x^{\tilde{k}_x}
d \boldsymbol{\alpha}},
\end{equation*}
where $\tilde{k}_x=\alpha_{0,x}+k_x-1$. In the special case of $\alpha_{0,x}=1$, $\forall x \in \mathcal{X}$, we have $\tilde{k}_x=k_x$.

The posterior average for $\alpha_{x'}$ is obtained directly from \eqref{eq:anymoment}, by choosing $k_x=\delta_{x=x'}$, i.e., $k_x=1$ for $x=x'$ and $k_x=0$ for $\forall x\neq x'$, while the posterior average for the product of $\alpha_{x'}$ and $\alpha_{x''}$ is obtained directly from \eqref{eq:anymoment},  by choosing $k_x=\delta_{x=x'}+\delta_{x=x''}$, i.e., $k_x=1$ for $x\in \left\{x', x''\right\}$ and $k_x=0$ for $\forall x\nin \left\{x', x''\right\}$.
\end{proof}

\begin{proof}[Proof of Theorem \ref{thm:theta_mean}]
Given $\boldsymbol{\alpha}$, the marginal posterior density for $\boldsymbol{\theta}_{X|y}$ is a Dirichlet distribution with parameters $s\boldsymbol{\alpha}+\boldsymbol{n}_y$, where $\boldsymbol{n}_y=(n_{x_1 y}, \ldots, n_{x_r y})'$.
It is thus easy to compute 
\begin{equation*}
\mean{\left.\theta_{x|y}\right| 
\boldsymbol{\alpha},D
}=\E^{
\boldsymbol{\alpha},D}\left[\theta_{x|y} \right]=\frac{n_{xy}+s\alpha_x}{n_y+s}, 
\end{equation*}
where $x \in \mathcal{X}$, $y \in \mathcal{Y}$, and $\E^{\boldsymbol{\alpha},D}\left[\cdot\right]= \E\left[\cdot \left| \boldsymbol{\alpha},D\right.\right]$.

The posterior average of $\theta_{x|y}$ can thus be computed by means of the law of total expectation as
\begin{equation*}
\E^{
D}\left[{\theta_{x|y}}\right]=\E^{
D}
\left[\E^{
\boldsymbol{\alpha},D}\left[{\theta_{x|y}}\right]\right]=\frac{n_{xy}+s\E^{
D}\left[{\alpha_x}\right]}{n_y+s}.
\end{equation*}
The posterior average of $\alpha_{x}$ is obtained directly from Lemma \ref{thm:moments}.

In order to compute the posterior covariance between $\theta_{x|y}$ and  $\theta_{x'|y'}$ we can use the law of total covariance, i.e.,
\begin{align*}
\Cov^{
D}{\left(\theta_{x|y}, \theta_{x'|y'}\right)}= &\ \Cov^{
D}\left(\E^{
\boldsymbol{\alpha},D}\left[\theta_{x|y}\right], \E^{
\boldsymbol{\alpha},D}\left[\theta_{x'|y'}\right]\right)+\\
&+\E^{s,D} \left[\Cov^{
\boldsymbol{\alpha},D}\left(\theta_{x|y}, \theta_{x'|y'}\right)\right].
\end{align*}
The first quantity is:
\begin{align*}
&\Cov^{D}\left(\E^{ \boldsymbol{\alpha},D}\left[\theta_{x|y}\right], \E^{ \boldsymbol{\alpha},D}\left[\theta_{x'|y'}\right]\right)\\
&=\Cov^{D}\left(\frac{n_{xy}+s\alpha_x}{n_y+s}, 
\frac{n_{x' y'}+s\alpha_{x'}}{n_{y'}+s}\right)=\frac{s^2 \Cov^{D}\left(\alpha_x,\alpha_{x'}\right)}{\left(n_y+s\right) \left(n_{y'}+s\right)}.
\end{align*}
If $y'=y$, the second quantity is:
\begin{align*}
 &\E^{D} \left[\Cov^{\boldsymbol{\alpha},D}\left(\theta_{x|y}, \theta_{x'|y}\right)\right]\\
 &=\E^{D} \left[\frac{(n_{xy}+s\alpha_x)((n_y+s)\delta_{x x'}-(n_{x' y}+s\alpha_{x'}))}{(n_y+s)^2(n_y+s+1)} \right]\\
  &=\frac{(n_{xy}+s\E^{D} \left[\alpha_x\right])\delta_{x x'}}{(n_y+s)(n_y+s+1)}\!- \!\frac{\E^{D} \left[(n_{xy}+s\alpha_x)(n_{x' y}+s\alpha_{x'})\right]}{(n_y+s)^2(n_y+s+1)} \\
& =\frac{\hat\theta_{x|y}\delta_{x x'}- \hat\theta_{x|y} \hat{\theta}_{x'|y'}}{n_y+s+1}\!-\! \frac{s^2\!\left(\E^{D}\! \left[\alpha_x \alpha_{x'}\right]\!-\!\E^{D}\! \left[\alpha_x\right]\E^D\!\left[ \alpha_{x'}\right]\right)}{(n_y+s)^2(n_y+s+1)}\\
& =\frac{\hat\theta_{x|y}\delta_{x x'}- \hat\theta_{x|y} \hat{\theta}_{x'|y'}}{n_y+s+1}- \frac{s^2\Cov^{D} \left(\alpha_x, \alpha_{x'}\right)}{(n_y+s)^2(n_y+s+1)}.
\end{align*}
Otherwise, if $y' \neq y$,  $\E^{D} \left[\Cov^{\boldsymbol{\alpha},D}\left(\theta_{x|y}, \theta_{x'|y'}\right)\right]=0$, since $\boldsymbol{\theta}_{X|y}\independent \boldsymbol{\theta}_{X|y'}$ given $\boldsymbol{\alpha}$.

Exploiting the law of total covariance, we obtain
\begin{align*}
&\Cov^{D}{\left(\theta_{x|y}, \theta_{x|y'}\right)}= \frac{s^2 \Cov^{D}\left(\alpha_x,\alpha_{x'}\right)}{\left(n_y+s\right) \left(n_{y'}+s\right)} + \nonumber\\
&+\delta_{y y'} \left(\frac{\hat\theta_{x|y}\delta_{x x'}- \hat\theta_{x|y}\hat{\theta}_{x'|y}}{n_y+s+1}- \frac{s^2\Cov^{D} \left(\alpha_x, \alpha_{x'}\right)}{(n_y+s)^2(n_y+s+1)} \right).
\end{align*}
The posterior covariance between $\alpha_{x}$ and $\alpha_{x'}$ is obtained directly from Lemma \ref{thm:moments}.
\end{proof}
%

\begin{proof}[Proof of Theorem \ref{prop:MSE_sh}]
Exploiting the linearity of the ideal estimator we obtain that
\begin{align*}
\theta_{x|y}^{*}- \theta_{x|y} 
= \tilde{\omega}_y \left(\theta_{x|y}^{\text{ML}}-\theta_{x|y} \right)  + (1-\tilde{\omega}_y)\left(\tilde{\alpha}_x -\theta_{x|y} \right).
\end{align*}
The MSE of $\theta_{x|y}^{*}$ is thus
\begin{align*}
&\text{MSE}\left(\theta_{x|y}^{*} \right)= \tilde{\omega}_y^2 \text{MSE}\left(\theta_{x|y}^{\text{ML}} \right) + (1-\tilde{\omega}_y)^2 \text{MSE}\left(\tilde{\alpha}_x \right)+\\
&+\tilde{\omega}_y (1-\tilde{\omega}_y)\E_{\theta}\left[\E_{n}\left[\left(\theta_{x|y}^{\text{ML}}-{\theta}_{x|y} \right)\left(\tilde{\alpha}_x-{\theta}_{x|y} \right)\right]\right].
\end{align*}

Using the definition of MSE \eqref{eq:MSE} and the assumptions of model \eqref{eq:hier_generative}, i.e., $\E_{n}\left[n_{xy}\right]=n_y \theta_{x|y}$, $\E_{\theta}\left[ \theta_{x|y} \right]=\tilde{\alpha}_x$ and $\Var_{\theta} \left(  \theta_{x|y}\right)= \frac{\tilde{\alpha}_x-\tilde{\alpha}_x^2}{s+1}$, we obtain
\begin{align}
& \text{MSE}\left(\theta_{x|y}^{\text{ML}} \right)  
 =\E_{\theta}\left[\E_{n}\left[\left(\frac{n_{xy}}{n_y}-{\theta}_{x|y} \right)^2\right]\right]\nonumber \\
 &=\E_{\theta}\left[\frac{1}{n_y^2}\Var_{n}\left(n_{xy} \right)\right] =\frac{n_y}{n_y^2}\E_{\theta}\left[ \theta_{x|y}\left(1-\theta_{x|y}\right) \right]\nonumber \\
 &=\frac{1}{n_y}\left(\E_{\theta}\left[ \theta_{x|y} \right]-\E_{\theta}\left[ \theta_{x|y} \right]^2 - \Var_{\theta} \left(  \theta_{x|y}\right) \right)\nonumber \\
 &= \frac{1}{n_y}\left( \tilde{\alpha}_x - \tilde{\alpha}_x^2 - \frac{\tilde{\alpha}_x-\tilde{\alpha}_x^2}{s+1}\right)= \frac{s}{n_y}\frac{\tilde{\alpha}_x-\tilde{\alpha}_x^2}{s+1}.
 \label{eq:MSE_ML_proof}
\end{align}
This quantity corresponds to the first term of $\text{MSE}\big(\theta_{x|y}^{*} \big)$. 
The second term is obtained as
\begin{align*}
\text{MSE}\left(\tilde{\alpha}_x \right)&=\E_{\theta}\left[\E_{n}\left[\left(\tilde{\alpha}_x-{\theta}_{x|y} \right)^2\right]\right]=\E_{\theta}\left[\left(\tilde{\alpha}_x-{\theta}_{x|y} \right)^2\right]\\
&=\Var_{\theta}\left(\theta_{x|y}\right)=\frac{\tilde{\alpha}_x-\tilde{\alpha}_x^2}{s+1},
\end{align*} 
since $\tilde{\alpha}_x-{\theta}_{x|y}$ is independent of $n_{xy}$ and $\E_{\theta}\left[\theta_{x|y}\right]=\tilde{\alpha}_x$. 
The last term $\E_{\theta}\big[\E_{n}\big[\big(\theta_{x|y}^{\text{ML}}-{\theta}_{x|y} \big)\big(\tilde{\alpha}_x-{\theta}_{x|y} \big)\big]\big]=0$, since $\tilde{\alpha}_x-{\theta}_{x|y}$ is independent of $n_{xy}$ and $\E_{\theta}\big[\theta_{x|y}^{\text{ML}}\big]=\theta_{x|y}$.\\
The MSE for the ideal estimator is thus
\begin{align*}
\text{MSE}\left(\theta_{x|y}^{*} \right)&= \tilde{\omega}_y^2  \frac{s}{n_y}\frac{\tilde{\alpha}_x-\tilde{\alpha}_x^2}{s+1}
 + (1-\tilde{\omega}_y)^2 \frac{\tilde{\alpha}_x-\tilde{\alpha}_x^2}{s+1}\\
 &= \left(\tilde{\omega}_y^2  \frac{s}{n_y}
 + (1-\tilde{\omega}_y)^2 \right) \frac{\tilde{\alpha}_x-\tilde{\alpha}_x^2}{s+1}.
\end{align*}
If $\tilde{\omega}_y=\frac{n_y}{n_y+s}$ and $s>0$, $\forall x \in \mathcal{X}$,
\begin{align*}
&\text{MSE}\left(\theta_{x|y}^{*} \right)=  \left(\frac{n_y^2}{(n_y+s)^2}  \frac{s}{n_y}
 + \frac{s^2}{(n_y+s)^2} \right) \frac{\tilde{\alpha}_x-\tilde{\alpha}_x^2}{s+1}\\
& 
= \frac{s }{n_y+s} \frac{\tilde{\alpha}_x-\tilde{\alpha}_x^2}{s+1}
< \frac{s}{n_y} \frac{\tilde{\alpha}_x-\tilde{\alpha}_x^2}{s+1} = \text{MSE}\left(\theta_{x|y}^{\text{ML}} \right).
\end{align*}
Thus, $\sum_{x\in\mathcal{X}}\text{MSE}\big(\theta_{x|y}^{*} \big) < \sum_{x\in\mathcal{X}}\text{MSE}\big(\theta_{x|y}^{\text{ML}} \big)$.

The MSE for the Bayesian estimator \eqref{eq:bayes} is:
\begin{align*}
&\text{MSE}\left(\theta_{x|y}^{\text{B}} \right)=\omega_y^2\text{MSE}\left(\theta_{x|y}^{\text{ML}} \right)+(1-\omega_y^2)\text{MSE}\left(\frac{1}{r}\right)+\\
&+\omega_y (1-\omega_y)\E_{\theta}\left[\E_{n}\left[\left(\theta_{x|y}^{\text{ML}}-{\theta}_{x|y} \right)\left(\frac{1}{r}-\theta_{x|y} \right)\right]\right].
\end{align*}
The first term is derived in \eqref{eq:MSE_ML_proof}.\\ 
The second term corresponds to
\begin{align*}
&\text{MSE}\left(\frac{1}{r}\right)=\E_{\theta}\left[\left(\frac{1}{r}-\tilde{\alpha}_x \right)^2\right]+\E_{\theta}\left[\left(\tilde{\alpha}_x-{\theta}_{x|y} \right)^2\right]\\
&=\left(\tilde{\alpha}_x-\frac{1}{r}\right)^2+ \Var_{\theta}\left(\theta_{x|y}\right)=\left(\tilde{\alpha}_x-\frac{1}{r}\right)^2+ \frac{\tilde{\alpha}_x-\tilde{\alpha}_x^2}{s+1},
\end{align*}
since $\frac{1}{r}-\theta_{x|y}$ is independent of $n_{xy}$ and $\E_{\theta}\left[\theta_{x|y}\right]=\tilde{\alpha}_x$. \\
The last term $\E_{\theta}\left[\E_{n}\left[\left(\theta_{x|y}^{\text{ML}}-{\theta}_{x|y} \right)\left(\frac{1}{r}-\theta_{x|y} \right)\right]\right]=0$, since $\frac{1}{r}-{\theta}_{x|y}$ is independent of $n_{xy}$ and $\E_{\theta}\left[\theta_{x|y}^{\text{ML}}\right]=\theta_{x|y}$.

The MSE for the Bayesian estimator is thus
\begin{align*}
&\text{MSE}\left(\theta_{x|y}^{\text{B}} \right)=\left(\omega_y^2  \frac{s}{n_y}
 + (1-\omega_y)^2 \right) \frac{\tilde{\alpha}_x-\tilde{\alpha}_x^2}{s+1}+\\
&+ (1-\omega_y)^2\left(\tilde{\alpha}_x-\frac{1}{r}\right)^2\!\!\!\!=\text{MSE}\left(\theta_{x|y}^{*} \right)\!+\! (1-\omega_y)^2\left(\tilde{\alpha}_x-\frac{1}{r}\right)^2\!\!\!,
\end{align*}
if $\tilde{\omega}_y=\omega_y$. Thus, $\text{MSE}\big(\theta_{x|y}^{*} \big)\leq \text{MSE}\big(\theta_{x|y}^{\text{B}} \big)$, $\forall x \in \mathcal{X} $. The two estimators have the same MSE in the special case of $\tilde{\alpha}_x=\frac{1}{r}$. As a consequence, $\sum_{x\in \mathcal{X}}\text{MSE}\big(\theta_{x|y}^{*} \big)\leq \sum_{x\in \mathcal{X}}\text{MSE}\big(\theta_{x|y}^{\text{B}} \big)$, with equality if $\tilde{\boldsymbol{\alpha}}=\frac{1}{r}\cdot \mathbf{1}_{1\times r}$.

In the finite sample assumption $\text{MSE}\left(\hat{\theta}_{x|y}\right)$ differs from $\text{MSE}\big({\theta}^{*}_{x|y}\big)$, since $\hat\theta_{x|y}$ includes 
an estimator of $\boldsymbol{\alpha}$. In particular, 
\begin{align}
&\text{MSE}\left(\hat{\theta}_{x|y}\right)=\text{MSE}\left(\theta_{x|y}^{*} \right)+(1-\omega_y)^2\text{MSE}\left(\hat{\alpha}_x \right)+\nonumber \\
&+\omega_y(1-\omega_y)\E_{\theta}\left[\E_{n}\left[\left({{\theta}}_{x|y}^{\text{ML}}-{\theta}_{x|y} \right)\left(\hat{\alpha}_x-\tilde{\alpha}_x \right)\right]\right].
\label{eq:MSE_hier}
\end{align}
The second and third term in \eqref{eq:MSE_hier} cannot be computed analytically and should be computed numerically. 
\end{proof}

\section*{Acknowledgment}
The research in this paper has been partially supported by the Swiss NSF grants ns.~IZKSZ2\_162188.
\bibliographystyle{IEEEtran}
\bibliography{biblio}


%

\end{document}